\documentclass[11pt,a4paper,logo,nonumbering]{aura}
\usepackage{xspace}
\usepackage{natbib}
\usepackage{booktabs}
\usepackage{float}
\usepackage{tabularx}
\usepackage{array}
\usepackage{pifont}
\usepackage{fontawesome5} 
\usepackage{multirow} 
\usepackage{xcolor}
\usepackage{tikz}
\IfFileExists{CJKutf8.sty}{\usepackage{CJKutf8}}{}
\hypersetup{colorlinks=true,allcolors=blue,pdftitle={Beyond Answering: Learning to Speak First and Act in Streaming Video}}
\newcommand{\ourmodel}{AURA 2.0\xspace}

\newcommand{\yes}{\ding{51}}
\newcommand{\no}{\ding{55}}

\newcolumntype{Y}{>{\centering\arraybackslash}X}
\title{Beyond Answering:\\ Learning to Speak First and Act in Streaming Video}

\newcommand{\AURAGitHubURL}{https://github.com/aurateam2026/Aura2.0}
\newcommand{\AURAHuggingFaceURL}{https://huggingface.co/aurateam/AURA_2}
\newcommand{\AURAHomepageURL}{https://aurateam2026.github.io/Aura-2.0/}

\author{Shuhan Li$^{1,*}$, Huankang Guan$^{1,*}$, Xintong Guo$^{1,*}$, Yang Bo$^{1,*,\dagger}$, Yuyang He$^{1,*}$, Dunyuan Xu$^{1,*}$, Jingyan Chen$^{1,*}$,
Fang Liu$^{1}$, Tsz Kin Wan$^{1}$, Jinpeng Chen$^{1,\S}$, Zhenqi He$^{1,3}$, Yuxuan Hu$^{1,2}$, \newline 
Wenbo Li$^{1,2}$, Han Gao$^{1}$, Qijun Li$^{1}$, Long Chen$^{3}$, Hongsheng Li$^{2}$, Rui Liu$^{1,\ddagger}$ \\\normalfont\normalsize $^{1}$Huawei Research \qquad $^{2}$CUHK MMLab \qquad $^{3}$HKUST\\\normalfont\normalsize $^{*}$Equal Contribution \qquad $^{\dagger}$Project Lead \qquad $^{\ddagger}$Corresponding Author}

\newcommand{\AURAProjectLinks}{\normalfont\normalsize
  \href{\AURAGitHubURL}{\textcolor{black}{\faGithub\hspace{0.35em}\textbf{GitHub}}}\hspace{1.5em}%
  \href{\AURAHuggingFaceURL}{\textcolor{black}{\raisebox{-0.15em}{\includegraphics[height=1.1em]{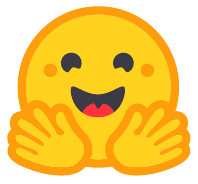}}\hspace{0.35em}\textbf{Hugging Face}}}\hspace{1.5em}%
  \href{\AURAHomepageURL}{\textcolor{black}{\faHome\hspace{0.35em}\textbf{Homepage}}}%
}
\date{}

\begin{document}
\begin{abstract}

Most AI assistants respond only when prompted, remaining silent even when events in the video stream offer meaningful opportunities for conversation. To move beyond purely user-driven interaction, we present \textbf{AURA 2.0}, a streaming video large language model that introduces three core upgrades to AURA. The key advance is \textbf{Model-Initiated Conversation (MIC)}, which enables the model to identify topics worth discussing in the evolving visual stream and initiate conversations at appropriate moments without a user query. \textbf{Stream-Aware Tool Calling (SATC)} lets the model act on what it sees by invoking external tools using the current visual and conversational context and incorporating returned information into the ongoing interaction. \textbf{Interactive Visual Grounding (IVG)} supports visual grounding in both user inputs and model outputs. The model can identify objects marked by users on video frames and produce frame-aligned bounding boxes to link its responses to specific objects and regions. To develop these capabilities, we construct a dedicated streaming interaction dataset and adopt a three-stage training scheme to integrate the new behaviors with AURA's existing streaming interaction abilities. We further incorporate \textbf{Long-Horizon Interaction Memory (LHIM)} to maintain contextual continuity over extended video streams. Evaluations on public streaming benchmarks show that AURA 2.0 achieves leading performance among open-source streaming models, while targeted evaluations further validate its new interaction capabilities.

\end{abstract}

\maketitle
\begingroup
\renewcommand{\thefootnote}{\S}
\footnotetext[0]{The author is no longer affiliated with the primary affiliation.}
\endgroup
\begin{figure}[h]
    \centering
    \includegraphics[width=0.8\linewidth]{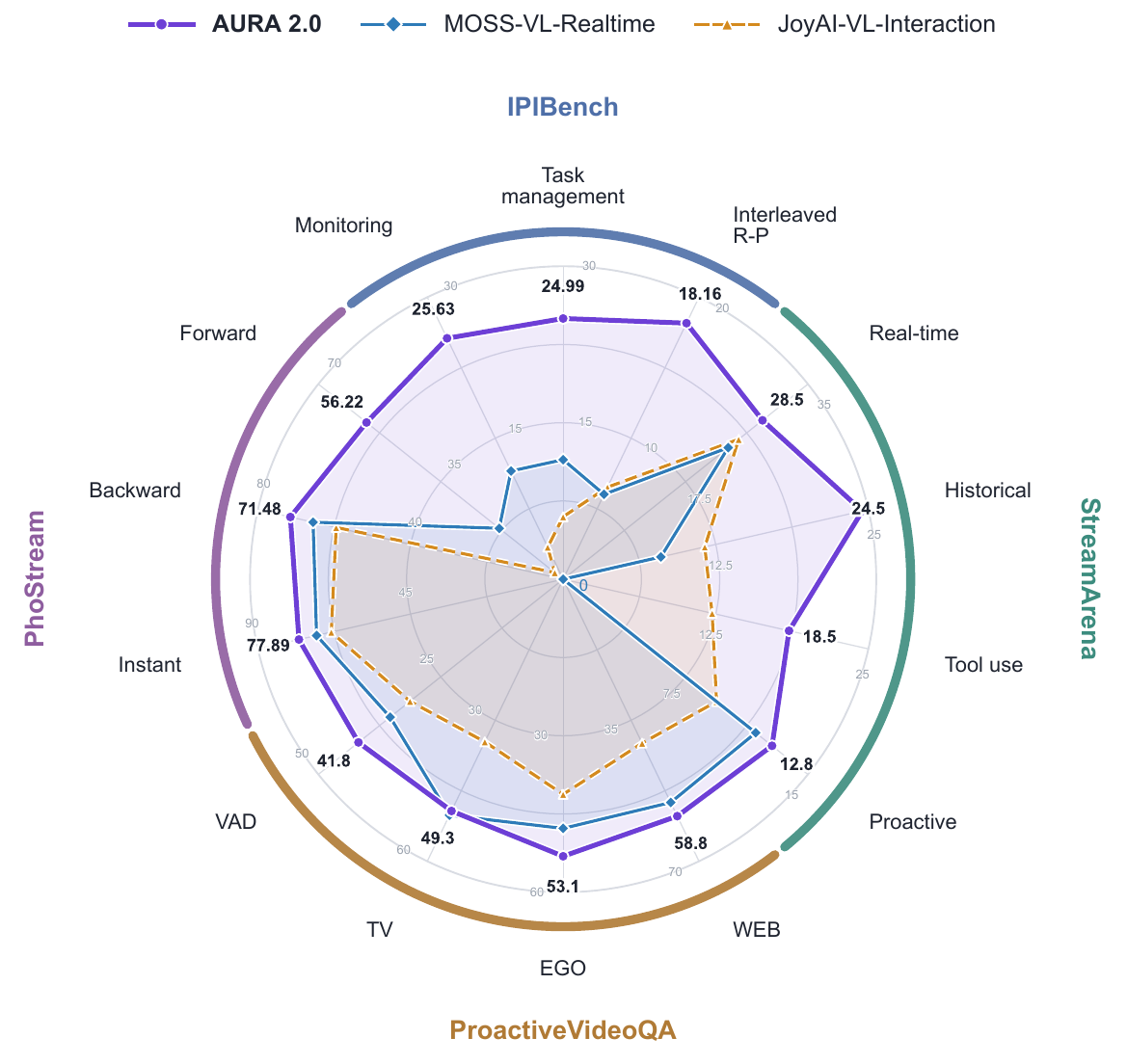}
    \caption{\textbf{Comparison of AURA 2.0 with MOSS-VL-Realtime and JoyAI-VL-Interaction across four public streaming benchmarks.} The Tool use score for MOSS-VL-Realtime on StreamArena is plotted as zero because the model does not support tool calling.
    }
    \label{fig:teaser}
\end{figure}

\section{Introduction}\label{sec:intro}
Streaming Video Large Language Models (VideoLLMs) extend video understanding from offline analysis to online interaction over video streams. Unlike offline models that analyze complete videos, streaming VideoLLMs process incoming frames incrementally and generate responses using only the information available so far~\citep{chen2024videollm,lu2026aura,yao2026harnessing}. Their designs coordinate continuous visual perception with response decisions, enabling models to remain silent during observation and respond when relevant evidence becomes available~\citep{qian2025dispider,wang2025mmduet2,zheng2026emgarde,yang2026livestarpro}. Context management and memory mechanisms further maintain contextual continuity over extended video streams without retaining all previous frames~\citep{lu2026aura,yang2026livestarpro,yao2026harnessing,zeng2026onestreamer}.
Despite these advances, interaction remains largely user-driven. Responses are triggered by user queries, with models responding either immediately or later when relevant evidence becomes available. However, events in the video stream may offer meaningful opportunities for conversation even without a user query.

To move beyond purely user-driven interaction, we introduce \textbf{Model-Initiated Conversation (MIC)}, which enables the model to identify topics worth discussing in the evolving visual stream and initiate conversations at appropriate moments without a user query, speaking first rather than only answering. In addition to deciding when to respond to a user query, the model must decide whether to initiate a conversation and what to discuss. These decisions depend on both the evolving visual stream and the conversational context: a visual event may offer a meaningful opportunity for conversation, but discussing it may be unnecessary if the topic has already been addressed. The model must therefore use this context to avoid repetitive or unnecessary interruptions.

Building on AURA~\citep{lu2026aura}, we present \textbf{AURA 2.0}, a streaming VideoLLM that introduces MIC as the key advance, alongside \textbf{Stream-Aware Tool Calling (SATC)} and \textbf{Interactive Visual Grounding (IVG)}, while retaining Real-Time QA, Proactive QA, and Multi-Response QA. SATC enables the model to act on what it sees by invoking external tools using the current visual and conversational context, answering questions beyond its own knowledge. Returned information is then incorporated into the ongoing interaction. IVG supports visual grounding in both user inputs and model outputs, allowing users and the model to specify which objects and regions they refer to. Users can mark objects on video frames for the model to identify, and the model can produce frame-aligned bounding boxes to link its responses to specific objects and regions.

To develop these capabilities, we construct a dedicated streaming interaction dataset and adopt a three-stage training scheme to integrate the new behaviors with AURA's existing streaming interaction abilities. Building on AURA's streaming interaction dataset, we introduce new data covering MIC, SATC, and IVG. The resulting dataset comprises approximately 11.4 billion tokens, spanning 5,400 hours of video and 40 million training frames. We employ multimodal large language model (MLLM)-based sample-level verification to ensure data quality and align queries, responses, tool calls, and bounding boxes to video chunks along a consistent timeline. During training, we apply focal-loss-based supervision to the response-control tokens to mitigate their imbalance and support appropriate response decisions.

Further, we incorporate \textbf{Long-Horizon Interaction Memory (LHIM)} and optimize the streaming inference framework to support ongoing interaction over extended video streams. As visual observations move beyond the active visual window, salient events are summarized into textual descriptions and incorporated into the ongoing conversational context. LHIM complements the existing visual context management mechanism of AURA by preserving information about events beyond the active visual window, enabling the model to refer to previously observed events when answering user queries or identifying topics worth discussing.

As shown in Figure~\ref{fig:teaser}, AURA 2.0 demonstrates strong performance on public streaming benchmarks, including PhoStream~\citep{lu2026phostream}, ProactiveVideoQA~\citep{wang2025proactivevideoqa}, IPIBench~\citep{li2026ipibench}, and StreamArena~\citep{zhang2026streamarena}. We further evaluate conversation initiation and response quality for MIC, tool-assisted question answering for SATC, and localization and abstention for IVG.

The main contributions of our work are summarized as follows.

\begin{itemize}

\item We introduce \textbf{AURA 2.0}, a streaming VideoLLM with MIC as the key advance, enabling the model to identify topics worth discussing and initiate conversations at appropriate moments without a user query. SATC and IVG further extend its interaction capabilities while retaining Real-Time QA, Proactive QA, and Multi-Response QA.

\item We construct a dedicated streaming interaction dataset covering MIC, SATC, and IVG by extending the streaming interaction dataset of AURA. The resulting dataset comprises approximately 11.4 billion tokens, 5,400 hours of video, and 40 million training frames. We adopt a three-stage training scheme with focal-loss-based supervision for response-control tokens to integrate the new behaviors with existing streaming interaction abilities.

\item We incorporate LHIM to maintain contextual continuity over extended video streams by preserving salient events as textual descriptions beyond the active visual window. We further optimize the streaming inference framework for low-latency deployment and timely interactions.

\item We demonstrate strong performance on public streaming benchmarks and validate AURA 2.0’s new interaction capabilities through targeted evaluations.

\end{itemize}

\section{Related Work}\label{sec:related}

\subsection{Streaming Video LLMs}

While offline VideoLLMs analyze complete video clips for comprehension, temporal reasoning, and task execution~\citep{team2024gemini,qwen2026qwen38,google2026gemini38flash,bytedance2026seed21}, streaming VideoLLMs process incoming observations incrementally and decide when to speak or remain silent using only the video observed so far. Early streaming works~\citep{chen2024videollm,zhang2024flash,chen2025livecc,fu2025vispeak,yao2025timechat,qian2025dispider,wang2025streambridge} introduce online perception and timely responses to incrementally arriving video. Within this setting, most systems remain \emph{query-driven}: the user specifies a question or monitoring goal, and the model primarily decides \emph{when} sufficient visual evidence has arrived to answer. AURA~\citep{lu2026aura} unifies Real-Time QA, Proactive QA, and Multi-Response QA to support immediate, delayed, and multiple answers within one streaming framework. MMDuet2~\citep{wang2025mmduet2} learns speaking and silence decisions through multi-turn reinforcement learning, while Em-Garde~\citep{zheng2026emgarde} triggers answers when observations match visual conditions derived from a user query.

Recent models further couple continuous perception with finer interaction control to support broader proactive behaviors such as standing instructions, alerts, live commentary, and scene-triggered reminders~\citep{yan2026proactvl}. JoyAI-VL-Interaction~\citep{yao2026joyai} makes a learned decision each second to remain silent, respond, or delegate across time-aligned QA, alerting, and commentary using explicit silence/response tokens. MOSS-VL-Realtime~\citep{wang2026mossvl} uses gated cross-attention and per-frame silence/response tokens to incorporate newly arriving frames, allowing it to wait, continue speaking, or revise its response under standing instructions and commentary. MiniCPM-o 4.5~\citep{cui2026minicpmo45} and SeedRealtime~\citep{bytedance2026seedrealtime} align audiovisual and text streams end-to-end to regulate speaking timing and deliver proactive reminders or suggestions, while Qwen3.8-Omni (via Qwen-Live-Harness)~\citep{qwen2026qwen38omni} and Gemini Live~\citep{google2026geminilive,google2026geminiproactivity,google2026gemini38live} additionally support user-defined monitoring, proactive response withholding, and continuous audiovisual dialogue~\citep{qwen2025qwen25omni,qwen2025qwen3omni,hurst2024gpt}. Streaming benchmarks assess complementary aspects of online understanding and proactive behavior~\citep{lin2024streamingbench,niu2025ovo,wang2025omnimmi,wang2025proactivevideoqa}. However, across these settings, responses remain conditioned on prior user questions, standing instructions, or predefined monitoring rules, and, to our knowledge, no existing model is trained to initiate open-ended conversations without any prior request. AURA 2.0 bridges this gap by introducing \textbf{Model-Initiated Conversation}: rather than merely deciding when to respond to a user request, the model proactively discovers topics worth discussing from the evolving visual scene and conversational history, determining both what to discuss and when to initiate a conversation while remaining silent when appropriate.

\begin{table*}[t]
\centering
\small
\setlength{\tabcolsep}{3pt}
\renewcommand{\arraystretch}{1.0}
\caption{\textbf{Comparison of representative streaming interaction models.} A check mark (\yes) indicates a publicly reported capability, while a cross (\no) indicates that supporting evidence was not identified, rather than a confirmed lack of capability.}
\label{tab:interaction_capabilities}
\def\AURACapabilityOriginal{\begin{tabular}{@{}lcccccc@{}}
\toprule
\textbf{Model} & \textbf{Real-Time QA} & \textbf{Proactive QA} & \textbf{Multi-Response QA} & \textbf{MIC} & \textbf{SATC} & \textbf{IVG} \\
\midrule
AURA~\citep{lu2026aura} & \yes & \yes & \yes & \no & \no & \no \\
MMDuet2~\citep{wang2025mmduet2} & \yes & \yes & \yes & \no & \no & \no \\
MOSS-VL-Realtime~\citep{wang2026mossvl} & \yes & \yes & \yes & \no & \no & \no \\
JoyAI-VL-Interaction~\citep{yao2026joyai} & \yes & \yes & \yes & \no & \yes & \no \\
MiniCPM-o~4.5~\citep{cui2026minicpmo45} & \yes & \yes & \yes & \no & \no & \no \\
\midrule
\textbf{AURA~2.0 (Ours)} & \yes & \yes & \yes & \yes & \yes & \yes \\
\bottomrule
\end{tabular}}
\sbox0{\AURACapabilityOriginal}
\edef\AURACapabilityTableWidth{\the\wd0}
\sbox0{\begin{tabular}{@{}l@{}}AURA~\citep{lu2026aura}\\MMDuet2~\citep{wang2025mmduet2}\\MOSS-VL-Realtime~\citep{wang2026mossvl}\\JoyAI-VL-Interaction~\citep{yao2026joyai}\\MiniCPM-o~4.5~\citep{cui2026minicpmo45}\\\textbf{AURA~2.0 (Ours)}\end{tabular}}
\edef\AURACapabilityModelWidth{\the\wd0}
\dimen2=\AURACapabilityTableWidth\relax
\advance\dimen2 by -\wd0
\advance\dimen2 by -36pt
\divide\dimen2 by 6
\sbox0{\textbf{Multi-Response}}
\dimen4=0.85\wd0
\advance\dimen4 by 2pt
\ifdim\dimen2<\dimen4 \dimen2=\dimen4\fi
\edef\AURACapabilityColWidth{\the\dimen2}
\resizebox{\textwidth}{!}{\begin{tabular}{@{}lcccccc@{}}
\toprule
\textbf{Model} & \makebox[\AURACapabilityColWidth][c]{\begin{tabular}[c]{@{}c@{}}\textbf{Real-Time}\\\textbf{QA}\end{tabular}} & \makebox[\AURACapabilityColWidth][c]{\begin{tabular}[c]{@{}c@{}}\textbf{Proactive}\\\textbf{QA}\end{tabular}} & \makebox[\AURACapabilityColWidth][c]{\begin{tabular}[c]{@{}c@{}}\textbf{Multi-Response}\\\textbf{QA}\end{tabular}} & \makebox[\AURACapabilityColWidth][c]{\textbf{MIC}} & \makebox[\AURACapabilityColWidth][c]{\textbf{SATC}} & \makebox[\AURACapabilityColWidth][c]{\textbf{IVG}} \\
\midrule
AURA~\citep{lu2026aura} & \makebox[\AURACapabilityColWidth][c]{\yes} & \makebox[\AURACapabilityColWidth][c]{\yes} & \makebox[\AURACapabilityColWidth][c]{\yes} & \makebox[\AURACapabilityColWidth][c]{\no} & \makebox[\AURACapabilityColWidth][c]{\no} & \makebox[\AURACapabilityColWidth][c]{\no} \\
MMDuet2~\citep{wang2025mmduet2} & \makebox[\AURACapabilityColWidth][c]{\yes} & \makebox[\AURACapabilityColWidth][c]{\yes} & \makebox[\AURACapabilityColWidth][c]{\yes} & \makebox[\AURACapabilityColWidth][c]{\no} & \makebox[\AURACapabilityColWidth][c]{\no} & \makebox[\AURACapabilityColWidth][c]{\no} \\
MOSS-VL-Realtime~\citep{wang2026mossvl} & \makebox[\AURACapabilityColWidth][c]{\yes} & \makebox[\AURACapabilityColWidth][c]{\yes} & \makebox[\AURACapabilityColWidth][c]{\yes} & \makebox[\AURACapabilityColWidth][c]{\no} & \makebox[\AURACapabilityColWidth][c]{\no} & \makebox[\AURACapabilityColWidth][c]{\no} \\
JoyAI-VL-Interaction~\citep{yao2026joyai} & \makebox[\AURACapabilityColWidth][c]{\yes} & \makebox[\AURACapabilityColWidth][c]{\yes} & \makebox[\AURACapabilityColWidth][c]{\yes} & \makebox[\AURACapabilityColWidth][c]{\no} & \makebox[\AURACapabilityColWidth][c]{\yes} & \makebox[\AURACapabilityColWidth][c]{\no} \\
MiniCPM-o~4.5~\citep{cui2026minicpmo45} & \makebox[\AURACapabilityColWidth][c]{\yes} & \makebox[\AURACapabilityColWidth][c]{\yes} & \makebox[\AURACapabilityColWidth][c]{\yes} & \makebox[\AURACapabilityColWidth][c]{\no} & \makebox[\AURACapabilityColWidth][c]{\no} & \makebox[\AURACapabilityColWidth][c]{\no} \\
\midrule
\textbf{AURA~2.0 (Ours)} & \makebox[\AURACapabilityColWidth][c]{\yes} & \makebox[\AURACapabilityColWidth][c]{\yes} & \makebox[\AURACapabilityColWidth][c]{\yes} & \makebox[\AURACapabilityColWidth][c]{\yes} & \makebox[\AURACapabilityColWidth][c]{\yes} & \makebox[\AURACapabilityColWidth][c]{\yes} \\
\bottomrule
\end{tabular}}
\end{table*}

\subsection{Streaming Tool Calling and Visual Grounding}

Tool use extends streaming interaction beyond language responses by enabling models to access external information and execute tasks. JoyAI-VL-Interaction~\citep{yao2026joyai} supports asynchronous task delegation to background models or agents while maintaining foreground interaction. Qwen-Live-Harness~\citep{qwen2026qwen38omni} connects real-time interaction with external agent backends, allowing conversations to continue during task execution. Gemini 3.8 Live~\citep{google2026gemini38live} supports non-blocking function calls and configurable handling of returned results. SeedRealtime~\citep{bytedance2026seedrealtime} incorporates tool use into live audiovisual interaction, while TML's Interaction Models~\citep{thinkingmachines2026interactionmodels} explore the coordination of foreground interaction and background reasoning. These works demonstrate different approaches to integrating external capabilities into continuous interaction. Building on this direction, AURA 2.0 introduces \textbf{Stream-Aware Tool Calling}, which enables the model to determine when to invoke external tools, select appropriate tools, and formulate arguments based on evolving visual observations and conversational context. Tool results are then incorporated into subsequent responses, allowing external information to support ongoing streaming interactions.

Visual grounding connects language references to specific objects and regions in the visual scene. Existing vision-language models support object and region localization in images and videos~\citep{bai2025qwen25vl,bai2025qwen3,wang2025internvl3,yang2025kwai}. In streaming interactions, spatial grounding requires identifying and localizing relevant objects as new visual observations continuously arrive. AURA 2.0 introduces \textbf{Interactive Visual Grounding}, enabling the model to interpret user-provided visual prompts and generate frame-aligned bounding boxes for objects and regions referenced in its responses. This allows the model to connect ongoing conversations with specific visual elements in the observed scene. 

Table~\ref{tab:interaction_capabilities} summarizes the publicly reported interaction capabilities of representative streaming models. AURA 2.0 retains Real-Time QA, Proactive QA, and Multi-Response QA while integrating MIC, SATC, and IVG within a unified streaming interaction framework.
\section{AURA 2.0 Framework}
\label{sec:aura_framework}

\subsection{Overall Architecture}

\begin{figure*}[t]
    \centering
    \includegraphics[width=\textwidth]{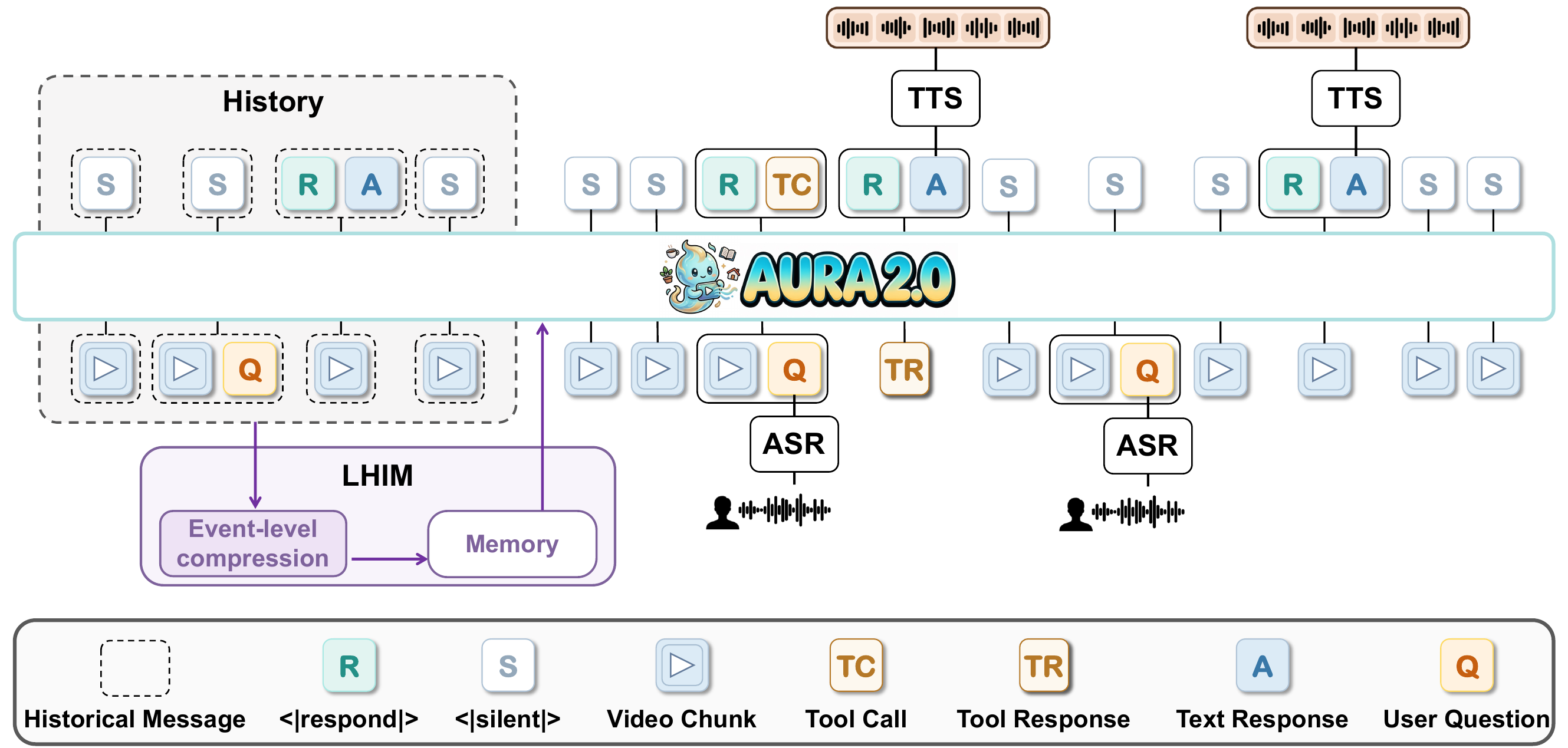}
    \caption{
        \textbf{Overview of the AURA 2.0 streaming interaction framework.} The model incrementally processes video chunks together with user inputs and interaction history, producing either \texttt{<|silent|>} for silent observation or \texttt{<|respond|>} followed by a text response or tool call.
Tool results are incorporated into the context to support subsequent responses.
Dashed outlines indicate historical messages beyond the active visual window.
    }
    \label{fig:overall_framework}
\end{figure*}

AURA 2.0 largely follows the streaming architecture and context management mechanism of AURA~\citep{lu2026aura}, as illustrated in Figure~\ref{fig:overall_framework}. Incoming video is organized into temporal chunks and processed incrementally together with the available user inputs and interaction history. The context retains recent visual observations and textual interactions through sliding windows, allowing the model to continuously observe the video stream and determine whether to generate a response.

For each video chunk, the assistant outputs \texttt{<|silent|>} alone when it remains silent. When it decides to respond, it outputs \texttt{<|respond|>} followed by response content, which can contain a conversational utterance or a tool call. This format explicitly distinguishes silent observation from response generation and accommodates different interaction behaviors within the same chunk-wise structure.

The main architectural addition is the LHIM module, implemented using an auxiliary summarization model. This model processes visual content leaving the active visual window together with all textual context available at the time of summarization. It organizes the information at the event level and produces summaries in QA form. These QA pairs are inserted back into the interaction context, preserving information about past events after their visual observations are no longer available. By incorporating both visual and textual information, LHIM provides historical context for subsequent responses and conversation initiation. Its implementation and integration into streaming inference are described in Section~\ref{sec:memory}.

\subsection{Streaming Interaction Capabilities}

The chunk-wise interaction format supports both user-driven responses and model-initiated conversations. AURA 2.0 retains Real-Time QA, Proactive QA, and Multi-Response QA from AURA, while extending its capabilities with MIC, SATC, and IVG.

\textbf{MIC}
enables the model to identify topics worth discussing in the evolving visual stream and initiate conversations when no new user query is present. The model determines whether an event warrants discussion, when to speak, and what to say based on both visual observations and the preceding conversational context. Previously discussed content informs these decisions to avoid repetitive or unnecessary interruptions. MIC is naturally accommodated by the existing interaction format. The user-text field for the corresponding chunk is empty, while the assistant generates \texttt{<|respond|>} followed by its utterance. An empty user-text field alone does not distinguish MIC from a deferred response. MIC initiates a conversational turn on the model's own initiative rather than answering a pending user query or carrying out a previously requested reminder.

\textbf{SATC}
enables the model to invoke external tools using the current visual and conversational context to obtain information beyond its own knowledge. Tool calls are represented as response content following \texttt{<|respond|>}, so the same response-control mechanism governs both conversational utterances and tool invocations. Returned information is incorporated into the ongoing interaction to support subsequent responses.

\textbf{IVG}
supports visual reference in both user inputs and model outputs. Users can mark objects or regions in video frames to specify what they refer to, and the model can associate its responses with visual targets by outputting frame-aligned bounding boxes. Each box is expressed in the standard coordinate format $[x_1, y_1, x_2, y_2]$ within the response content. This representation allows localization information to be conveyed through the same output format used for other responses.
\section{Data Construction}
\label{sec:data_construction}

\subsection{Data Construction Pipeline}
\label{subsec:data_pipeline}

\begin{figure*}[t]
    \centering
    \includegraphics[width=\textwidth]{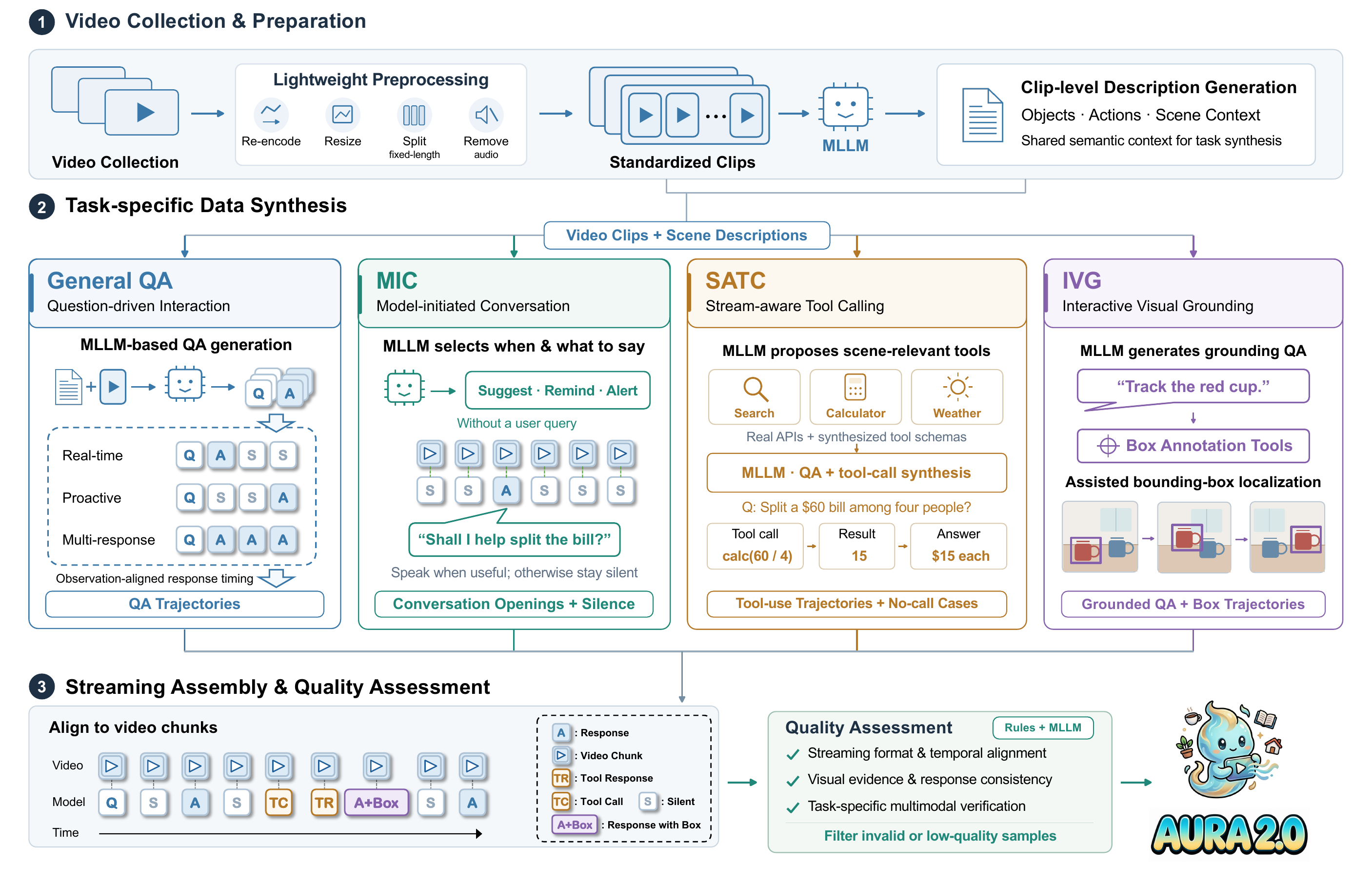}
    \caption{
        \textbf{Data construction pipeline for AURA 2.0.}
        The pipeline transforms collected videos into a verified streaming training mixture through three stages:
        (1) Video collection and preparation standardizes videos into clips and generates clip-level descriptions.
        (2) Task-specific data synthesis uses the clips and descriptions to construct General QA, MIC, SATC, and IVG trajectories.
        (3) Streaming assembly and quality assessment aligns the trajectories to video chunks and filters invalid or low-quality samples.
    }
    \label{fig:data_pipeline}
\end{figure*}

As shown in Figure~\ref{fig:data_pipeline}, the AURA 2.0 data pipeline extends AURA's streaming interaction data with three new forms of supervision: MIC, SATC, and IVG. It proceeds from video preparation to task-specific synthesis and finally to streaming assembly and quality verification. Source videos are re-encoded, resized, and divided into clips; an MLLM generates clip-level descriptions of objects, actions, and scene context. These descriptions support the construction of the inherited streaming QA tasks and the three new interaction types. The synthesized trajectories are aligned to video chunks so that queries, silent observations, responses, tool calls, and bounding boxes are represented along a consistent timeline.

\noindent\textbf{MIC data.} We construct MIC training data by asking an MLLM to determine when to initiate a conversation and what to say. For each video, the MLLM generates zero or more conversation openings, each consisting of a timestamp and a short message. Each opening must be supported by visual observations available up to that timestamp. When preceding dialogue is available from the data-construction process, the MLLM also uses it to select topics relevant to the user’s expressed interests and avoid repeating earlier discussions. Possible openings include drawing attention to a visual detail, offering a timely alert, providing contextual information, or suggesting a relevant activity. The prompt encourages useful contributions and discourages unsupported claims and unnecessary interruptions.

\noindent\textbf{SATC data.} We combine real execution records with MLLM-generated tool-calling traces to teach the model when and how to use external information. Each trajectory includes the tool-call arguments, the returned result, and the subsequent response. To encourage selective tool use, we add one to four randomly selected tool schemas to 50\% of non-tool-calling streaming samples while retaining their original targets. These examples teach the model to respond or remain silent without invoking a tool, even when tools are available.

\noindent\textbf{IVG data.}
We construct IVG examples from approximately 55,000 video grounding and tracking trajectories sourced from MeViS~\citep{ding2023mevis}, Refer-YouTube-VOS~\citep{seo2020urvos}, DanceTrack~\citep{sun2022dancetrack}, and SportsMOT~\citep{cui2023sportsmot}.
An MLLM rewrites the annotations into streaming grounding interactions, retaining available bounding-box coordinates and aligning them with their corresponding video frames.
The data cover six patterns: responses with boxes, multi-candidate disambiguation, subjective selection with an explanation, multi-turn follow-up and target switching, box-only tracking, and target-absent responses.

We apply structural checks and task-specific MLLM verification to filter synthesized trajectories. Structural checks ensure chronological consistency and correct alignment between bounding boxes and video frames. MLLM validators assess whether responses are supported by the available visual evidence and occur at appropriate times, using relevant video history for memory-related QA. MIC openings are reviewed for timing and scene consistency using clips extending five seconds beyond the proposed initiation time. For IVG, bounding boxes are rendered on the corresponding frames to verify target correctness. Samples that fail these checks are excluded from training.

\begin{figure}[t]
    \centering
    \includegraphics[width=\linewidth]{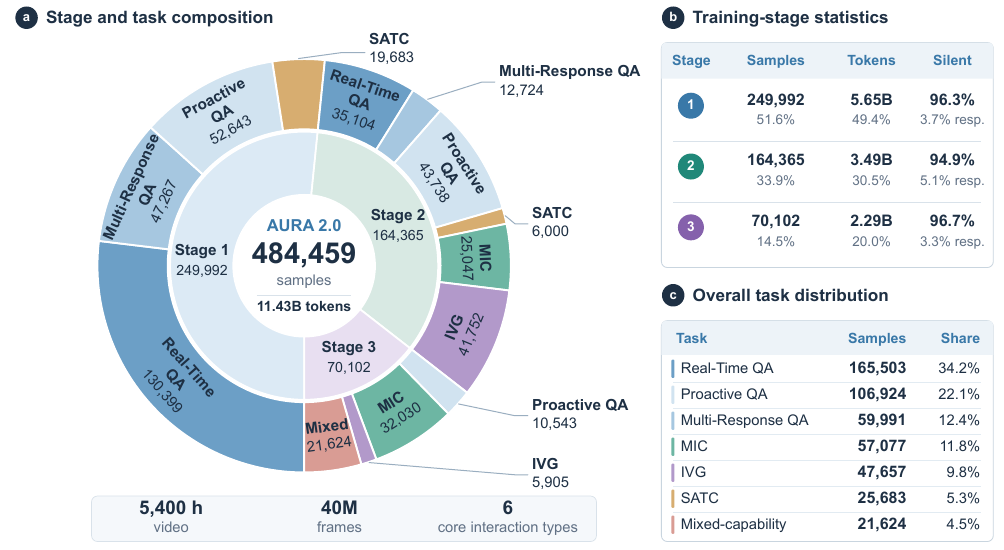}
    \caption{
        \textbf{Training data composition and response-control statistics across the three training stages.}
        (a) Sample distribution by stage (inner ring) and task (outer ring).
        (b) Stage-wise sample and token allocations and silent/response decision proportions.
        (c) Task-wise sample counts and shares aggregated across stages.
        Together, these distributions characterize the task coverage and supervision balance of the training mixture.
    }
    \label{fig:data_distribution}
\end{figure}

\subsection{Dataset Composition and Statistics}
\label{subsec:data_overview}

The verified training mixture retains Real-Time QA, Proactive QA, and Multi-Response QA alongside MIC, SATC, IVG, and mixed-capability data. Across the three training stages, it comprises 484,459 samples and 11.43 billion tokens, covering about 5,400 hours of video and 40 million training frames. The sample and token counts include replayed examples across stages and should not be interpreted as numbers of distinct source videos.

As shown in Figure~\ref{fig:data_distribution}, the inherited query-driven tasks remain the largest part of the sample distribution: Real-Time QA accounts for 34.2\%, Proactive QA for 22.1\%, and Multi-Response QA for 12.4\%. These samples provide supervision for immediate answers, delayed responses when sufficient evidence becomes available, and multiple responses to a single user query, respectively. The newly introduced data cover MIC (11.8\%), IVG (9.8\%), SATC (5.3\%), and mixed-capability examples (4.5\%). These data extend supervision to conversation initiation, visual reference, selective tool use, and capability composition, while retaining supervision for the existing QA tasks.

The three stages account for 51.6\%, 33.9\%, and 14.5\% of samples, compared with 49.4\%, 30.5\%, and 20.0\% of tokens, respectively. Stage 3 therefore has a greater token share than sample share, indicating a higher average token count per sample. The rationale for the stage-wise task mixtures is discussed in Section~\ref{sec:three_stage}. Silent observations dominate the response-control targets in all three stages (96.3\%, 94.9\%, and 96.7\%). This imbalance motivates the response-control objective in Section~\ref{sec:loss}.

\section{Training Strategy}
\label{sec:training}

AURA 2.0 is initialized from Qwen3.5-9B~\citep{qwen2026qwen35} and follows the streaming interaction architecture of AURA~\citep{lu2026aura}. Training must preserve AURA's existing response behavior while introducing MIC, SATC, and IVG. Because these capabilities change both response timing and output format, we introduce them progressively rather than learning all behaviors from a single mixture.

\subsection{Three-Stage Training Strategy}
\label{sec:three_stage}

Our three-stage curriculum first establishes general streaming interaction and tool use, then introduces MIC and IVG with replay of earlier tasks, and finally trains on examples that combine multiple capabilities. This progression aims to integrate new behaviors without degrading the existing response policy. Figure~\ref{fig:data_distribution} shows the stage-wise sample and token distributions; the token-level task ratios and their roles in each stage are detailed below. Each stage is trained for one epoch, with Stage~1 initialized from Qwen3.5-9B and subsequent stages initialized from the preceding stage.

\paragraph{Stage 1: streaming interaction and tool-use foundation.}
The first stage contains approximately 250k samples and 5.65B tokens, of which 87.2\% come from general streaming QA and 12.8\% from SATC. General streaming QA comprises Real-Time QA, Proactive QA, and Multi-Response QA, covering immediate, delayed, and continuous responses to user queries. Together, these data establish a basic response policy in which the model remains silent while observing, responds when sufficient evidence is available, and produces multiple updates when needed. SATC is included at this stage to enable the model to determine when and how to use tools based on user requests and the evolving visual and conversational context.

\paragraph{Stage 2: visual grounding and model-initiated conversation.}
The second stage contains approximately 164k samples and 3.49B tokens. Data retained from Stage~1 account for 65.3\% of the tokens, while MIC and IVG account for 25.7\% and 8.9\%, respectively. In the MIC data, model responses are aligned with relevant visual evidence to supervise both response timing and content. IVG introduces bounding-box tokens for visually grounded references. Retaining earlier data helps prevent the new tasks from increasing unnecessary responses, producing bounding boxes in ordinary answers, or degrading previously learned behaviors.

\paragraph{Stage 3: capability composition and preference refinement.}
The final stage contains approximately 70k samples and 2.29B tokens. Mixed-capability data account for 34.9\% of the tokens, MIC for 46.5\%, Proactive QA for 13.3\%, and IVG for 5.4\%. Examples that combine capabilities, such as Proactive QA with SATC and MIC with SATC, teach the model to coordinate response timing, tool use, and model-initiated interaction within the same trajectory. To further refine when the model should engage, some Proactive QA examples supervise silence when an event is irrelevant or a response would be repetitive.

\subsection{Response-Control Objective and Optimization}
\label{sec:loss}

As shown in Figure~\ref{fig:data_distribution}b, silent steps account for more than 94\% of the supervised control decisions in every stage. On average, each clip contains 24, 24, and 42 silent steps in Stages~1--3, respectively. This imbalance can cause standard token-level cross-entropy to overemphasize frequent silence decisions relative to less frequent response decisions. To address this issue, we combine class-balanced normalization
with focal weighting for the control tokens \texttt{<|silent|>} and \texttt{<|respond|>},
while using standard cross-entropy for all other supervised tokens.

Let $p_i$ denote the predicted probability of the ground-truth token at position $i$, and let $\ell_i=-\log p_i$. All probabilities are computed over the full vocabulary under next-token prediction. We use $S$ and $R$ to denote the sets of supervised \texttt{<|silent|>} and \texttt{<|respond|>} positions, respectively, and $O$ to denote all other supervised positions, including response text, serialized tool-call content, and end-of-turn sequences. These sets are defined over the global optimizer-step batch, including all accumulated microbatches and data-parallel workers. When both control classes and the other-token group are present, the training objective is
\begin{equation}
\mathcal{L} =
\frac{1}{4}\left(
\frac{1}{|S|}\sum_{i\in S}(1-p_i)^{\gamma}\,\ell_i
+
\frac{1}{|R|}\sum_{i\in R}(1-p_i)^{\gamma}\,\ell_i
\right)
+
\frac{1}{|O|}\sum_{j\in O}\ell_j,
\label{eq:control_loss}
\end{equation}
with $\gamma=2$.

The focal factor $(1-p_i)^{\gamma}$ is applied only to the two control classes and remains part of the gradient computation. It reduces the contribution of confidently predicted control tokens and places greater emphasis on difficult decisions, such as remaining silent when the model tends to respond or responding when it tends to remain silent. Response text and serialized tool-call content use standard cross-entropy without focal weighting.

We average the losses of the two control classes separately. Thus, silence and response receive equal base weights regardless of their respective numbers of supervised positions in the global batch. All other supervised tokens are averaged together using standard cross-entropy. We assign a weight of $0.25$ to each of the two class-level mean focal losses and a weight of $1.0$ to the mean cross-entropy over all other supervised tokens. These coefficients are applied directly and are not renormalized by their sum. We use the same loss weights in all three stages, without a separate response classifier or reinforcement learning.

\noindent\textbf{Optimization details.} \label{sec:optim} We perform full-parameter SFT of the language model with a global batch size of 32 sequences, a micro-batch size of 1, and no gradient accumulation. We use bin packing to limit each packed sequence to 65,536 tokens and apply padding-free attention. The visual encoder and vision--language aligner remain frozen. Training uses bf16, AdamW with $\beta_1=0.9$, $\beta_2=0.95$, a weight decay of 0.1, gradient clipping at 1.0, and activation recomputation. We use a cosine learning-rate schedule with 10\% warmup. The peak learning rate is $1\times10^{-5}$ for Stage~1 and $2\times10^{-6}$ for Stages~2 and 3; the corresponding minimum learning rates are $1\times10^{-6}$ and $1\times10^{-7}$. The three-stage training process requires approximately 1,100 accelerator-hours in total.

\section{Streaming Inference and Long-Horizon Memory}
\label{sec:inference_system}

AURA 2.0 largely inherits the asynchronous streaming inference pipeline of AURA~\citep{lu2026aura}, integrating incremental video processing, context management, Automatic Speech Recognition (ASR), and Text-to-Speech (TTS). It retains \texttt{<|silent|>} for silent observation and introduces \texttt{<|respond|>} before each non-silent response. Conversational text is passed to TTS, tool calls are executed with results returned to the model, and bounding boxes are rendered on the frontend. We further integrate LHIM to preserve historical information and optimize inference latency, as detailed in the following two subsections.

\subsection{Long-Horizon Interaction Memory}
\label{sec:memory}

To support long-horizon streaming video understanding without continuously increasing the visual context, AURA 2.0 employs LHIM, a hierarchical event memory that compresses historical video observations into timestamped textual descriptions. 
The context is organized into two components: a sliding window that preserves the most recent 30 seconds of video and the associated dialogue, and a context history that stores compressed visual memories alongside earlier question-answer interactions.
Video preceding the sliding window is divided into 30-second slices and processed by a separate summarization MLLM. 
Given uniformly sampled frames and the associated dialogue as reference, the MLLM partitions each slice into semantically coherent events. 
Each event contains start and end timestamps and a concise description of the salient visual information.
Event timestamps are converted to absolute session times. The events from each 30-second slice form a level-1 memory record, which is stored as a text-only QA pair in the context history.

As the session progresses, older records are progressively merged into coarser summaries. 
At each memory update, adjacent level-1 records are merged in pairs into level-2 summaries when these records end at least five minutes before the update time. 
Adjacent level-2 records are similarly merged into level-3 summaries after ten minutes. 
Each merged record preserves the combined time span and key visual information while reducing temporal detail. 
Level-3 records remain in the context rather than being discarded to free capacity, while dialogue within the window remains aligned with its corresponding video chunks.
Together, these mechanisms maintain a fixed 30-second visual context, preserve recent history at event resolution, and retain older observations as progressively more compact text tied to absolute session time.

\subsection{Inference Latency Optimization}\label{sec:latency_opt}

We serve \ourmodel with vLLM-Omni~\citep{vllmomni2026technicalreport}. Each turn is triggered either by a new video chunk or by a user query and is processed as a single request, which passes through four stages: speech recognition, MLLM inference, speech synthesis, and codec decoding (waveform generation). If a new turn begins with \texttt{<|respond|>} while the previous response is still being spoken, the ongoing response is interrupted and replaced by the new one; the full text of the interrupted response is retained in the dialogue history. Turns that output \texttt{<|silent|>} do not affect ongoing speech. We introduce three optimizations to this pipeline, and Section~\ref{sec:exp_efficiency} evaluates the contribution of each.

\noindent\textbf{Session history.} Each request carries only the frames of the current video chunk. The server decodes these frames into pixel arrays once and caches them, together with the dialogue, in a per-session history that mirrors the sliding window of Section~\ref{sec:memory}. When constructing the input of the next turn, the cached arrays are reused in their original order, so earlier frames are neither retransmitted by the client nor decoded again by the server. When context truncation removes video chunks from the sliding window, the corresponding arrays are evicted from the history, which keeps its size bounded by the window length. In addition, prefix caching reuses the KV cache of the unchanged context prefix across turns, so only newly appended frames and dialogue require prefill; the cache is rebuilt after each truncation.

\noindent\textbf{Fast paths for silent input and silent output.} For turns triggered by a video chunk without user speech, speech recognition is skipped, and an empty transcript is passed to the MLLM together with the frames. On the output side, the first generated token determines the path: a \texttt{<|silent|>} token ends the turn after a single decoding step, without running speech synthesis or codec decoding. Only turns that begin with \texttt{<|respond|>} proceed to speech synthesis. Tool calls and bounding boxes are removed from the text stream before it reaches speech synthesis, and a turn that contains only a tool call does not invoke speech synthesis.

\noindent\textbf{Sentence-level speech handoff.} Speech synthesis starts before the MLLM finishes generating the full response. Generated text is split at sentence and clause boundaries, and completed segments are forwarded to speech synthesis once they total at least 15 tokens; shorter segments are buffered and merged with the following ones. All text of a turn is appended to a single streaming speech synthesis request, which remains open while generation continues.

\section{Experiments}
\label{sec:experiments}

We evaluate AURA 2.0 from three perspectives: streaming interaction performance, newly introduced interaction capabilities, and inference efficiency. After describing the evaluation settings (Section~\ref{sec: exp_setup}), we report results on four public benchmarks that assess response timing, event understanding, proactive task management, historical reasoning, tool use, and the coordination of reactive and proactive interactions (Section~\ref{sec:streaming_bench_results}). We then conduct targeted evaluations of MIC, SATC, and IVG to assess model-initiated conversations, the contribution of external tools to answer accuracy, and visual localization, respectively (Sections~\ref{sec:MIC_results}--\ref{sec:eval_ivg}). Finally, we measure response latency under concurrent sessions and analyze the effects of the serving optimizations (Section~\ref{sec:exp_efficiency}).

\subsection{Experimental Setup} \label{sec: exp_setup}

AURA 2.0 processes incoming video incrementally in one-second chunks, with frame sampling governed by the benchmark-specific protocol below. The model maintains a 30-second sliding visual window together with textual interaction history and uses only information available up to the current step. User messages are inserted at their annotated timestamps, and previous user messages and generated responses are retained as textual context after their frames leave the visual window.

\noindent\textbf{IPIBench.} For JoyAI-VL-Interaction, MOSS-VL-Realtime, and AURA 2.0, we process videos at 1 FPS with a 30-second visual context window. The models decide whether to remain silent or respond at every step. Qwen3.5-9B and Qwen3.5-9B-thinking are evaluated separately using the official offline inference pipeline, with their visual context extended to 30 frames. We follow the official scoring protocol and report subtask accuracies together with the arithmetic mean of the subtask scores within each category.

\noindent\textbf{StreamArena.} 
JoyAI-VL-Interaction (with memory module), MOSS-VL-Realtime, and AURA 2.0 (with LHIM) process videos at 1 FPS with a 30-second visual context window. RTP, HR, and Pro are evaluated together in a single run, with dialogue history preserved across these tasks. Tool is evaluated separately because it requires external network access, and its dialogue history is not shared with the other tasks. Outputs are converted into a unified response format and scored using the official StreamArena protocol. We report accuracy for each evaluation dimension and its corresponding subcategories.

\noindent\textbf{ProactiveVideoQA.} JoyAI-VL-Interaction and AURA 2.0 process videos at 2 FPS with a 30-second visual context window. Observations outside the window are compressed into textual memory to preserve historical information. We follow the official scoring protocol and report PAUC separately for WEB, EGO, TV, and VAD at $\omega\in\{0,0.5,1\}$. The results for MOSS-VL-Realtime are taken from its official technical report.

\noindent\textbf{PhoStream.} Qwen3.5-9B, JoyAI-VL-Interaction, MOSS-VL-Realtime, and AURA 2.0 process videos at 2 FPS with a 30-second visual context window.
Following PhoStream's official inference schedule, we run inference once at the question timestamp for Instant and Backward tasks, and at offsets of 0, 1, and 2 seconds from both the question and ground-truth answer timestamps for Forward tasks.
We follow the official response-window and scoring protocol, using gpt-4.1 as the LLM-based judge.

\subsection{Performance on Public Streaming Benchmarks} \label{sec:streaming_bench_results}

\subsubsection{Evaluation on IPIBench}

\def\AURAComparisonFontReference{\begin{tabular}{@{}lc*{7}{c}@{}}
        \toprule
        \multirow{2}{*}{\textbf{Model}}
        & \multirow{2}{*}{\textbf{Param}}
        & \multicolumn{3}{c}{\textbf{Evaluation Score} ($\uparrow$)}
        & \multirow{2}{*}{\textbf{Overall} ($\uparrow$)}
        & \multicolumn{3}{c}{\textbf{Forward Task Analysis (\%)}}
        \\
        \cmidrule(lr){3-5}
        \cmidrule(l){7-9}
        & &
        Instant
        & Backward
        & Forward
        &
        & ER ($\downarrow$)
        & NR ($\downarrow$)
        & PC ($\uparrow$)
        \\
        \midrule

        \multicolumn{9}{@{}l}{
            \textit{Proprietary Offline Models}
        } \\

        Gemini 3 Pro
        & --
        & 80.83 & 82.19 & 16.40
        & 48.70
        & 79.12 & 0.11 & 20.77
        \\

        Doubao-Seed-1.6
        & --
        & 71.28 & 62.94 & 44.26
        & 56.01
        & 29.76 & 13.38 & 56.85
        \\

        Doubao-Seed-1.8
        & --
        & 80.45 & 77.31 & 33.38
        & 56.15
        & 56.46 & 2.36 & 41.18
        \\

        \midrule
        \multicolumn{9}{@{}l}{
            \textit{Open-source Offline Models}
        } \\

        Qwen2.5-Omni-7B
        & 7B
        & 67.71 & 65.20 & 1.81
        & 34.06
        & 42.65 & 43.50 & 13.85
        \\

        Qwen3-VL-8B
        & 8B
        & 75.22 & 71.50 & 7.18
        & 40.25
        & 85.44 & 3.50 & 11.06
        \\

        Qwen3.5-9B
        & 9B
        & 73.29 & 63.26 & 11.18
        & 39.86
        & 66.52 & 17.70 & 15.77
        \\

        Qwen3-VL-30B-A3B
        & 30B
        & 73.38 & 69.46 & 5.25
        & 38.33
        & 91.33 & 1.18 & 7.49
        \\

        Qwen3-Omni-30B-A3B
        & 30B
        & 77.18 & \textbf{77.24} & 1.26
        & 39.02
        & 97.89 & \textbf{0.07} & 2.03
        \\

        \midrule
        \multicolumn{9}{@{}l}{
            \textit{Open-source Streaming Models}
        } \\

        Dispider
        & 7B
        & 44.24 & 42.90 & 3.53
        & 23.50
        & 68.52 & 21.20 & 10.28
        \\

        VideoLLM-online-8B
        & 8B
        & 24.88 & 24.72 & 0.00
        & 12.34
        & 99.54 & 0.43 & 0.04
        \\

        MMDuet2
        & 3B
        & 8.76 & 8.30 & 1.39
        & 4.96
        & 28.55 & 59.21 & 12.24
        \\

        JoyAI-VL-Interaction
        & 8B
        & 68.39 & 59.43 & 2.40
        & 33.21
        & 94.93 & 1.32 & 3.75
        \\

        MOSS-VL-Realtime
        & 11B
        & 73.86 & 65.89 & 18.27
        & 44.30
        & 41.51 & 28.87 & 29.62
        \\
        \midrule
        \textbf{AURA 2.0 (Ours)}
        & 9B
        & \textbf{77.89} & 71.48 & \textbf{56.22}
        & \textbf{65.69}
        & \textbf{1.30} & 23.90 & \textbf{74.80}
        \\

        \bottomrule
    \end{tabular}}

\begin{table*}[!t]
    \centering
    \caption{\textbf{Main results on IPIBench.}
    Results for previously evaluated models are reproduced from the original IPIBench paper, while results for AURA 2.0, JoyAI-VL-Interaction, MOSS-VL-Realtime, and Qwen3.5-9B (with thinking disabled and enabled) are obtained from our evaluations.
    All scores are reported as accuracies (\%), and Avg.\ denotes the arithmetic mean of the three subtask scores within each category.
    Boldface indicates the best result among open-source streaming models in each metric column.}
    \label{tab:ipibench_main_results}

    \setlength{\tabcolsep}{4pt}
    \renewcommand{\arraystretch}{1.12}

\sbox0{\AURAComparisonFontReference}
\edef\AURATargetTableWidth{\the\wd0}
\setlength{\tabcolsep}{2pt}
\sbox0{\begin{tabular}{@{}l@{}}Qwen3.5-9B-thinking\\LLaVA-OneVision-7B\\JoyAI-VL-Interaction\\MOSS-VL-Realtime\end{tabular}}
\dimen2=\AURATargetTableWidth\relax
\advance\dimen2 by -\wd0
\advance\dimen2 by -48pt
\divide\dimen2 by 12
\edef\AURAIPIMetricWidth{\the\dimen2}
\resizebox{\textwidth}{!}{%
\begin{tabular*}{\AURATargetTableWidth}{@{}l*{12}{c}@{}}
        \toprule

        \multirow{2}{*}{\textbf{Model}}
        & \multicolumn{4}{c}{\shortstack{\textbf{Proactive}\\\textbf{Monitoring}}}
        & \multicolumn{4}{c}{\shortstack{\textbf{Proactive Task}\\\textbf{Management}}}
        & \multicolumn{4}{c}{\shortstack{\textbf{Interleaved}\\\textbf{Reactive--Proactive}}}
        \\

        \cmidrule(lr){2-5}
        \cmidrule(lr){6-9}
        \cmidrule(l){10-13}

        & \makebox[\AURAIPIMetricWidth][c]{\small Timing} & \makebox[\AURAIPIMetricWidth][c]{\small Under.} & \makebox[\AURAIPIMetricWidth][c]{\small Repeat.} & \makebox[\AURAIPIMetricWidth][c]{\small Avg.}
        & \makebox[\AURAIPIMetricWidth][c]{\small Cancel} & \makebox[\AURAIPIMetricWidth][c]{\small Modify} & \makebox[\AURAIPIMetricWidth][c]{\small Multi} & \makebox[\AURAIPIMetricWidth][c]{\small Avg.}
        & \makebox[\AURAIPIMetricWidth][c]{\small R2P} & \makebox[\AURAIPIMetricWidth][c]{\small RuP} & \makebox[\AURAIPIMetricWidth][c]{\small RaP} & \makebox[\AURAIPIMetricWidth][c]{\small Avg.}
        \\

        \midrule

        Human Level
        & \makebox[\AURAIPIMetricWidth][c]{98.00} & \makebox[\AURAIPIMetricWidth][c]{94.00} & \makebox[\AURAIPIMetricWidth][c]{92.00} & \makebox[\AURAIPIMetricWidth][c]{94.67}
        & \makebox[\AURAIPIMetricWidth][c]{98.00} & \makebox[\AURAIPIMetricWidth][c]{92.00} & \makebox[\AURAIPIMetricWidth][c]{98.00} & \makebox[\AURAIPIMetricWidth][c]{96.00}
        & \makebox[\AURAIPIMetricWidth][c]{96.00} & \makebox[\AURAIPIMetricWidth][c]{96.00} & \makebox[\AURAIPIMetricWidth][c]{98.00} & \makebox[\AURAIPIMetricWidth][c]{96.67}
        \\

        \midrule
        \multicolumn{13}{@{}l}{\textit{Proprietary Offline Models}}
        \\

        Gemini 3 Pro
        & \makebox[\AURAIPIMetricWidth][c]{43.49} & \makebox[\AURAIPIMetricWidth][c]{24.90} & \makebox[\AURAIPIMetricWidth][c]{18.40} & \makebox[\AURAIPIMetricWidth][c]{28.93}
        & \makebox[\AURAIPIMetricWidth][c]{19.44} & \makebox[\AURAIPIMetricWidth][c]{21.84} & \makebox[\AURAIPIMetricWidth][c]{19.39} & \makebox[\AURAIPIMetricWidth][c]{20.22}
        & \makebox[\AURAIPIMetricWidth][c]{24.76} & \makebox[\AURAIPIMetricWidth][c]{17.68} & \makebox[\AURAIPIMetricWidth][c]{26.69} & \makebox[\AURAIPIMetricWidth][c]{23.04}
        \\

        Gemini 2.5 Pro
        & \makebox[\AURAIPIMetricWidth][c]{45.36} & \makebox[\AURAIPIMetricWidth][c]{25.26} & \makebox[\AURAIPIMetricWidth][c]{13.21} & \makebox[\AURAIPIMetricWidth][c]{27.94}
        & \makebox[\AURAIPIMetricWidth][c]{24.54} & \makebox[\AURAIPIMetricWidth][c]{20.81} & \makebox[\AURAIPIMetricWidth][c]{16.78} & \makebox[\AURAIPIMetricWidth][c]{20.71}
        & \makebox[\AURAIPIMetricWidth][c]{21.17} & \makebox[\AURAIPIMetricWidth][c]{17.54} & \makebox[\AURAIPIMetricWidth][c]{28.24} & \makebox[\AURAIPIMetricWidth][c]{22.32}
        \\

        GPT-5.4
        & \makebox[\AURAIPIMetricWidth][c]{54.64} & \makebox[\AURAIPIMetricWidth][c]{23.52} & \makebox[\AURAIPIMetricWidth][c]{8.96} & \makebox[\AURAIPIMetricWidth][c]{29.04}
        & \makebox[\AURAIPIMetricWidth][c]{38.43} & \makebox[\AURAIPIMetricWidth][c]{24.50} & \makebox[\AURAIPIMetricWidth][c]{7.57} & \makebox[\AURAIPIMetricWidth][c]{23.50}
        & \makebox[\AURAIPIMetricWidth][c]{27.04} & \makebox[\AURAIPIMetricWidth][c]{15.06} & \makebox[\AURAIPIMetricWidth][c]{29.28} & \makebox[\AURAIPIMetricWidth][c]{23.79}
        \\

        GPT-4o
        & \makebox[\AURAIPIMetricWidth][c]{50.59} & \makebox[\AURAIPIMetricWidth][c]{25.99} & \makebox[\AURAIPIMetricWidth][c]{13.68} & \makebox[\AURAIPIMetricWidth][c]{30.09}
        & \makebox[\AURAIPIMetricWidth][c]{11.57} & \makebox[\AURAIPIMetricWidth][c]{31.88} & \makebox[\AURAIPIMetricWidth][c]{14.18} & \makebox[\AURAIPIMetricWidth][c]{19.21}
        & \makebox[\AURAIPIMetricWidth][c]{27.04} & \makebox[\AURAIPIMetricWidth][c]{17.68} & \makebox[\AURAIPIMetricWidth][c]{26.17} & \makebox[\AURAIPIMetricWidth][c]{23.63}
        \\

        Qwen3.5-Plus
        & \makebox[\AURAIPIMetricWidth][c]{35.25} & \makebox[\AURAIPIMetricWidth][c]{24.35} & \makebox[\AURAIPIMetricWidth][c]{7.89} & \makebox[\AURAIPIMetricWidth][c]{22.50}
        & \makebox[\AURAIPIMetricWidth][c]{12.96} & \makebox[\AURAIPIMetricWidth][c]{31.54} & \makebox[\AURAIPIMetricWidth][c]{10.87} & \makebox[\AURAIPIMetricWidth][c]{18.46}
        & \makebox[\AURAIPIMetricWidth][c]{32.25} & \makebox[\AURAIPIMetricWidth][c]{22.01} & \makebox[\AURAIPIMetricWidth][c]{30.57} & \makebox[\AURAIPIMetricWidth][c]{28.28}
        \\

        \midrule
        \multicolumn{13}{@{}l}{\textit{Open-source Offline Models}}
        \\

        LLaVA-OneVision-7B
        & \makebox[\AURAIPIMetricWidth][c]{12.14} & \makebox[\AURAIPIMetricWidth][c]{8.66} & \makebox[\AURAIPIMetricWidth][c]{0.00} & \makebox[\AURAIPIMetricWidth][c]{6.93}
        & \makebox[\AURAIPIMetricWidth][c]{1.39} & \makebox[\AURAIPIMetricWidth][c]{1.01} & \makebox[\AURAIPIMetricWidth][c]{0.47} & \makebox[\AURAIPIMetricWidth][c]{0.96}
        & \makebox[\AURAIPIMetricWidth][c]{3.91} & \makebox[\AURAIPIMetricWidth][c]{0.00} & \makebox[\AURAIPIMetricWidth][c]{3.89} & \makebox[\AURAIPIMetricWidth][c]{2.60}
        \\

        InternVL3-8B
        & \makebox[\AURAIPIMetricWidth][c]{32.51} & \makebox[\AURAIPIMetricWidth][c]{22.31} & \makebox[\AURAIPIMetricWidth][c]{3.77} & \makebox[\AURAIPIMetricWidth][c]{19.53}
        & \makebox[\AURAIPIMetricWidth][c]{6.94} & \makebox[\AURAIPIMetricWidth][c]{6.04} & \makebox[\AURAIPIMetricWidth][c]{3.55} & \makebox[\AURAIPIMetricWidth][c]{5.51}
        & \makebox[\AURAIPIMetricWidth][c]{7.17} & \makebox[\AURAIPIMetricWidth][c]{0.00} & \makebox[\AURAIPIMetricWidth][c]{9.33} & \makebox[\AURAIPIMetricWidth][c]{5.50}
        \\

        Qwen3-VL-8B
        & \makebox[\AURAIPIMetricWidth][c]{43.12} & \makebox[\AURAIPIMetricWidth][c]{24.51} & \makebox[\AURAIPIMetricWidth][c]{4.25} & \makebox[\AURAIPIMetricWidth][c]{23.96}
        & \makebox[\AURAIPIMetricWidth][c]{1.39} & \makebox[\AURAIPIMetricWidth][c]{17.45} & \makebox[\AURAIPIMetricWidth][c]{9.22} & \makebox[\AURAIPIMetricWidth][c]{9.35}
        & \makebox[\AURAIPIMetricWidth][c]{24.43} & \makebox[\AURAIPIMetricWidth][c]{19.75} & \makebox[\AURAIPIMetricWidth][c]{23.05} & \makebox[\AURAIPIMetricWidth][c]{22.41}
        \\

        GLM-4.6V
        & \makebox[\AURAIPIMetricWidth][c]{50.43} & \makebox[\AURAIPIMetricWidth][c]{27.59} & \makebox[\AURAIPIMetricWidth][c]{10.85} & \makebox[\AURAIPIMetricWidth][c]{29.62}
        & \makebox[\AURAIPIMetricWidth][c]{15.28} & \makebox[\AURAIPIMetricWidth][c]{24.16} & \makebox[\AURAIPIMetricWidth][c]{9.69} & \makebox[\AURAIPIMetricWidth][c]{16.38}
        & \makebox[\AURAIPIMetricWidth][c]{22.15} & \makebox[\AURAIPIMetricWidth][c]{6.77} & \makebox[\AURAIPIMetricWidth][c]{39.37} & \makebox[\AURAIPIMetricWidth][c]{22.76}
        \\

        Qwen3.5-9B
        & \makebox[\AURAIPIMetricWidth][c]{5.59}
        & \makebox[\AURAIPIMetricWidth][c]{4.94}
        & \makebox[\AURAIPIMetricWidth][c]{0.94}
        & \makebox[\AURAIPIMetricWidth][c]{3.82}
        & \makebox[\AURAIPIMetricWidth][c]{0.46}
        & \makebox[\AURAIPIMetricWidth][c]{7.38}
        & \makebox[\AURAIPIMetricWidth][c]{2.13}
        & \makebox[\AURAIPIMetricWidth][c]{3.32}
        & \makebox[\AURAIPIMetricWidth][c]{27.04}
        & \makebox[\AURAIPIMetricWidth][c]{14.92}
        & \makebox[\AURAIPIMetricWidth][c]{3.11}
        & \makebox[\AURAIPIMetricWidth][c]{15.02}
        \\

        Qwen3.5-9B-thinking
        & \makebox[\AURAIPIMetricWidth][c]{41.00}
        & \makebox[\AURAIPIMetricWidth][c]{25.73}
        & \makebox[\AURAIPIMetricWidth][c]{13.21}
        & \makebox[\AURAIPIMetricWidth][c]{26.65}
        & \makebox[\AURAIPIMetricWidth][c]{0.00}
        & \makebox[\AURAIPIMetricWidth][c]{25.17}
        & \makebox[\AURAIPIMetricWidth][c]{8.75}
        & \makebox[\AURAIPIMetricWidth][c]{11.31}
        & \makebox[\AURAIPIMetricWidth][c]{15.96}
        & \makebox[\AURAIPIMetricWidth][c]{10.36}
        & \makebox[\AURAIPIMetricWidth][c]{33.16}
        & \makebox[\AURAIPIMetricWidth][c]{19.83}
        \\

        \midrule
        \multicolumn{13}{@{}l}{\textit{Open-source Streaming Models}}
        \\

        VideoLLM-online-8B
        & \makebox[\AURAIPIMetricWidth][c]{14.63} & \makebox[\AURAIPIMetricWidth][c]{0.14} & \makebox[\AURAIPIMetricWidth][c]{1.82} & \makebox[\AURAIPIMetricWidth][c]{5.53}
        & \makebox[\AURAIPIMetricWidth][c]{7.87} & \makebox[\AURAIPIMetricWidth][c]{1.68} & \makebox[\AURAIPIMetricWidth][c]{6.62} & \makebox[\AURAIPIMetricWidth][c]{5.39}
        & \makebox[\AURAIPIMetricWidth][c]{0.00} & \makebox[\AURAIPIMetricWidth][c]{1.10} & \makebox[\AURAIPIMetricWidth][c]{1.55} & \makebox[\AURAIPIMetricWidth][c]{0.88}
        \\

        Dispider
        & \makebox[\AURAIPIMetricWidth][c]{18.72} & \makebox[\AURAIPIMetricWidth][c]{0.65} & \makebox[\AURAIPIMetricWidth][c]{1.82} & \makebox[\AURAIPIMetricWidth][c]{7.06}
        & \makebox[\AURAIPIMetricWidth][c]{--} & \makebox[\AURAIPIMetricWidth][c]{--} & \makebox[\AURAIPIMetricWidth][c]{--} & \makebox[\AURAIPIMetricWidth][c]{--}
        & \makebox[\AURAIPIMetricWidth][c]{--} & \makebox[\AURAIPIMetricWidth][c]{--} & \makebox[\AURAIPIMetricWidth][c]{--} & \makebox[\AURAIPIMetricWidth][c]{--}
        \\

        Flash-VStream-7B
        & \makebox[\AURAIPIMetricWidth][c]{5.91} & \makebox[\AURAIPIMetricWidth][c]{0.00} & \makebox[\AURAIPIMetricWidth][c]{0.00} & \makebox[\AURAIPIMetricWidth][c]{1.97}
        & \makebox[\AURAIPIMetricWidth][c]{2.31} & \makebox[\AURAIPIMetricWidth][c]{0.00} & \makebox[\AURAIPIMetricWidth][c]{0.00} & \makebox[\AURAIPIMetricWidth][c]{0.77}
        & \makebox[\AURAIPIMetricWidth][c]{0.00} & \makebox[\AURAIPIMetricWidth][c]{0.00} & \makebox[\AURAIPIMetricWidth][c]{0.78} & \makebox[\AURAIPIMetricWidth][c]{0.26}
        \\

        JoyAI-VL-Interaction
        & \makebox[\AURAIPIMetricWidth][c]{9.53}
        & \makebox[\AURAIPIMetricWidth][c]{0.22}
        & \makebox[\AURAIPIMetricWidth][c]{0.47}
        & \makebox[\AURAIPIMetricWidth][c]{3.41}
        & \makebox[\AURAIPIMetricWidth][c]{14.81}
        & \makebox[\AURAIPIMetricWidth][c]{2.68}
        & \makebox[\AURAIPIMetricWidth][c]{0.47}
        & \makebox[\AURAIPIMetricWidth][c]{5.99}
        & \makebox[\AURAIPIMetricWidth][c]{16.94}
        & \makebox[\AURAIPIMetricWidth][c]{0.83}
        & \makebox[\AURAIPIMetricWidth][c]{1.81}
        & \makebox[\AURAIPIMetricWidth][c]{6.53}
        \\

        MOSS-VL-Realtime
        & \makebox[\AURAIPIMetricWidth][c]{23.41}
        & \makebox[\AURAIPIMetricWidth][c]{2.15}
        & \makebox[\AURAIPIMetricWidth][c]{8.96}
        & \makebox[\AURAIPIMetricWidth][c]{11.51}
        & \makebox[\AURAIPIMetricWidth][c]{15.28}
        & \makebox[\AURAIPIMetricWidth][c]{9.40}
        & \makebox[\AURAIPIMetricWidth][c]{9.69}
        & \makebox[\AURAIPIMetricWidth][c]{11.46}
        & \makebox[\AURAIPIMetricWidth][c]{7.17}
        & \makebox[\AURAIPIMetricWidth][c]{0.00}
        & \makebox[\AURAIPIMetricWidth][c]{10.88}
        & \makebox[\AURAIPIMetricWidth][c]{6.02}
        \\

        \midrule
        \textbf{AURA 2.0 (Ours)}
        & \makebox[\AURAIPIMetricWidth][c]{\textbf{49.30}}
        & \makebox[\AURAIPIMetricWidth][c]{\textbf{9.68}}
        & \makebox[\AURAIPIMetricWidth][c]{\textbf{17.92}}
        & \makebox[\AURAIPIMetricWidth][c]{\textbf{25.63}}
        & \makebox[\AURAIPIMetricWidth][c]{\textbf{37.96}}
        & \makebox[\AURAIPIMetricWidth][c]{\textbf{19.13}}
        & \makebox[\AURAIPIMetricWidth][c]{\textbf{17.87}}
        & \makebox[\AURAIPIMetricWidth][c]{\textbf{24.99}}
        & \makebox[\AURAIPIMetricWidth][c]{\textbf{24.10}}
        & \makebox[\AURAIPIMetricWidth][c]{\textbf{11.74}}
        & \makebox[\AURAIPIMetricWidth][c]{\textbf{18.65}}
        & \makebox[\AURAIPIMetricWidth][c]{\textbf{18.16}}
        \\
        \bottomrule
    \end{tabular*}%
}
\end{table*}

In addition to the results reported in the original paper, we evaluate five model configurations: JoyAI-VL-Interaction, MOSS-VL-Realtime, AURA 2.0, Qwen3.5-9B, and Qwen3.5-9B-thinking.
All results are reported as accuracies (\%), following the official IPIBench evaluation protocol.
Proactive Monitoring measures the ability to identify appropriate intervention timing, understand relevant visual events, and consistently trigger repeated actions.
Proactive Task Management evaluates cancellation, modification, and simultaneous management of multiple proactive tasks.
Interleaved Reactive--Proactive Interaction assesses the ability to handle reactive queries before, during, or after ongoing proactive tasks.

As shown in Table~\ref{tab:ipibench_main_results}, AURA 2.0 demonstrates strong proactive interaction capabilities.
In Proactive Monitoring, AURA 2.0 achieves a Timing accuracy of 49.30\%, outperforming Qwen3.5-9B, Qwen3-VL-8B, and Gemini 3 Pro, while approaching GPT-4o (50.59\%).
It also achieves 17.92\% in Repeated Proactiveness, approaching the strongest published baseline, Gemini 3 Pro (18.40\%).
AURA 2.0 achieves the highest Proactive Task Management average among all evaluated models, including proprietary ones.
It achieves 37.96\% in Cancellation, approaching GPT-5.4 (38.43\%), and obtains a Multi-task Management accuracy of 17.87\%, exceeding Gemini 2.5 Pro and GPT-5.4.
These results suggest that AURA 2.0 is effective at maintaining multiple concurrent proactive objectives and suppressing obsolete actions when tasks are cancelled.
For Interleaved Reactive--Proactive Interaction, AURA 2.0 achieves 24.10\% on R2P, 11.74\% on RuP, and 18.65\% on RaP. Its R2P performance is comparable to that of Gemini 3 Pro (24.76\%).

\subsubsection{Evaluation on StreamArena}

\begin{table*}[!t]
\centering
\caption{\textbf{Main results on StreamArena.}
We report the performance of AURA 2.0 and representative baselines across four evaluation dimensions: RTP, HR, Tool, and Pro.
Results for previously evaluated models, including the human reference, are reproduced from the original StreamArena paper, while results for AURA 2.0, JoyAI-VL-Interaction, and MOSS-VL-Realtime are obtained from our evaluations. Human HR results are reported with video rewatching in the upper row and without rewatching in the lower row.
All scores are reported as accuracies (\%), with higher values indicating better performance.
Boldface indicates the best result among open-source streaming models in each metric column.
}
\begingroup
\setlength{\tabcolsep}{4pt}
\renewcommand{\arraystretch}{1.12}

\sbox0{\AURAComparisonFontReference}
\edef\AURATargetTableWidth{\the\wd0}
\setlength{\tabcolsep}{2pt}
\sbox0{\begin{tabular}{@{}l@{}}Qwen3.5-397B-A17B\\JoyAI-VL-Interaction\\MOSS-VL-Realtime\\\textbf{AURA 2.0 (Ours)}\end{tabular}}
\dimen2=\AURATargetTableWidth\relax
\advance\dimen2 by -\wd0
\advance\dimen2 by -56pt
\divide\dimen2 by 11
\edef\AURAStreamMetricWidth{\the\dimen2}
\resizebox{\textwidth}{!}{%
\begin{tabular*}{\AURATargetTableWidth}{@{}l@{\hspace{16pt}}*{11}{c}@{}}
\toprule
\multirow{2}{*}{\textbf{Model}}
& \makebox[\AURAStreamMetricWidth][c]{\multirow{2}{*}{\textbf{RTP}$\,\uparrow$}} & \multicolumn{5}{c}{\textbf{HR}$\,\uparrow$}
& \makebox[\AURAStreamMetricWidth][c]{\multirow{2}{*}{\textbf{Tool}$\,\uparrow$}} & \multicolumn{4}{c}{\textbf{Pro.}$\,\uparrow$} \\
\cmidrule(lr){3-7}
\cmidrule(l){9-12}
& \makebox[\AURAStreamMetricWidth][c]{} & \makebox[\AURAStreamMetricWidth][c]{Avg.} & \makebox[\AURAStreamMetricWidth][c]{L1} & \makebox[\AURAStreamMetricWidth][c]{L2} & \makebox[\AURAStreamMetricWidth][c]{L3} & \makebox[\AURAStreamMetricWidth][c]{L4} & \makebox[\AURAStreamMetricWidth][c]{} & \makebox[\AURAStreamMetricWidth][c]{Avg.} & \makebox[\AURAStreamMetricWidth][c]{L1} & \makebox[\AURAStreamMetricWidth][c]{L2} & \makebox[\AURAStreamMetricWidth][c]{L3} \\
\midrule

\multirow{2}{*}{Human}
& \makebox[\AURAStreamMetricWidth][c]{\multirow{2}{*}{91.8}} & \makebox[\AURAStreamMetricWidth][c]{80.7} & \makebox[\AURAStreamMetricWidth][c]{84.2} & \makebox[\AURAStreamMetricWidth][c]{82.1} & \makebox[\AURAStreamMetricWidth][c]{79.5} & \makebox[\AURAStreamMetricWidth][c]{75.8} & \makebox[\AURAStreamMetricWidth][c]{\multirow{2}{*}{95.2}} & \makebox[\AURAStreamMetricWidth][c]{\multirow{2}{*}{91.5}} & \makebox[\AURAStreamMetricWidth][c]{\multirow{2}{*}{94.6}} & \makebox[\AURAStreamMetricWidth][c]{\multirow{2}{*}{91.3}} & \makebox[\AURAStreamMetricWidth][c]{\multirow{2}{*}{83.4}} \\
& \makebox[\AURAStreamMetricWidth][c]{} & \makebox[\AURAStreamMetricWidth][c]{63.4} & \makebox[\AURAStreamMetricWidth][c]{78.6} & \makebox[\AURAStreamMetricWidth][c]{68.9} & \makebox[\AURAStreamMetricWidth][c]{56.4} & \makebox[\AURAStreamMetricWidth][c]{45.1} & \makebox[\AURAStreamMetricWidth][c]{} & \makebox[\AURAStreamMetricWidth][c]{} & \makebox[\AURAStreamMetricWidth][c]{} & \makebox[\AURAStreamMetricWidth][c]{} & \makebox[\AURAStreamMetricWidth][c]{} \\
\midrule

\multicolumn{12}{@{}l}{\textit{Proprietary Offline Models}} \\
Gemini 3.5 Flash
& \makebox[\AURAStreamMetricWidth][c]{51.3} & \makebox[\AURAStreamMetricWidth][c]{51.4} & \makebox[\AURAStreamMetricWidth][c]{45.8} & \makebox[\AURAStreamMetricWidth][c]{57.8} & \makebox[\AURAStreamMetricWidth][c]{51.4} & \makebox[\AURAStreamMetricWidth][c]{45.3} & \makebox[\AURAStreamMetricWidth][c]{70.8} & \makebox[\AURAStreamMetricWidth][c]{--} & \makebox[\AURAStreamMetricWidth][c]{--} & \makebox[\AURAStreamMetricWidth][c]{--} & \makebox[\AURAStreamMetricWidth][c]{--} \\

Qwen3.5-Omni
& \makebox[\AURAStreamMetricWidth][c]{41.8} & \makebox[\AURAStreamMetricWidth][c]{35.8} & \makebox[\AURAStreamMetricWidth][c]{31.5} & \makebox[\AURAStreamMetricWidth][c]{44.8} & \makebox[\AURAStreamMetricWidth][c]{34.1} & \makebox[\AURAStreamMetricWidth][c]{25.4} & \makebox[\AURAStreamMetricWidth][c]{49.4} & \makebox[\AURAStreamMetricWidth][c]{--} & \makebox[\AURAStreamMetricWidth][c]{--} & \makebox[\AURAStreamMetricWidth][c]{--} & \makebox[\AURAStreamMetricWidth][c]{--} \\
\midrule

\multicolumn{12}{@{}l}{\textit{Open-source Offline Models}} \\
Qwen3.5-397B-A17B
& \makebox[\AURAStreamMetricWidth][c]{44.1} & \makebox[\AURAStreamMetricWidth][c]{41.5} & \makebox[\AURAStreamMetricWidth][c]{44.2} & \makebox[\AURAStreamMetricWidth][c]{47.3} & \makebox[\AURAStreamMetricWidth][c]{39.3} & \makebox[\AURAStreamMetricWidth][c]{30.4} & \makebox[\AURAStreamMetricWidth][c]{62.2} & \makebox[\AURAStreamMetricWidth][c]{--} & \makebox[\AURAStreamMetricWidth][c]{--} & \makebox[\AURAStreamMetricWidth][c]{--} & \makebox[\AURAStreamMetricWidth][c]{--} \\

MiMo-V2.5
& \makebox[\AURAStreamMetricWidth][c]{38.0} & \makebox[\AURAStreamMetricWidth][c]{35.8} & \makebox[\AURAStreamMetricWidth][c]{29.3} & \makebox[\AURAStreamMetricWidth][c]{45.5} & \makebox[\AURAStreamMetricWidth][c]{31.9} & \makebox[\AURAStreamMetricWidth][c]{28.7} & \makebox[\AURAStreamMetricWidth][c]{47.9} & \makebox[\AURAStreamMetricWidth][c]{--} & \makebox[\AURAStreamMetricWidth][c]{--} & \makebox[\AURAStreamMetricWidth][c]{--} & \makebox[\AURAStreamMetricWidth][c]{--} \\

Kimi-K2.6
& \makebox[\AURAStreamMetricWidth][c]{47.9} & \makebox[\AURAStreamMetricWidth][c]{43.8} & \makebox[\AURAStreamMetricWidth][c]{42.5} & \makebox[\AURAStreamMetricWidth][c]{53.3} & \makebox[\AURAStreamMetricWidth][c]{38.4} & \makebox[\AURAStreamMetricWidth][c]{33.1} & \makebox[\AURAStreamMetricWidth][c]{60.9} & \makebox[\AURAStreamMetricWidth][c]{--} & \makebox[\AURAStreamMetricWidth][c]{--} & \makebox[\AURAStreamMetricWidth][c]{--} & \makebox[\AURAStreamMetricWidth][c]{--} \\
\midrule

\multicolumn{12}{@{}l}{\textit{Open-source Streaming Models}} \\
AURA
& \makebox[\AURAStreamMetricWidth][c]{28.1} & \makebox[\AURAStreamMetricWidth][c]{22.7} & \makebox[\AURAStreamMetricWidth][c]{\textbf{25.4}} & \makebox[\AURAStreamMetricWidth][c]{27.0} & \makebox[\AURAStreamMetricWidth][c]{\textbf{24.3}} & \makebox[\AURAStreamMetricWidth][c]{10.5} & \makebox[\AURAStreamMetricWidth][c]{--} & \makebox[\AURAStreamMetricWidth][c]{3.7} & \makebox[\AURAStreamMetricWidth][c]{8.9} & \makebox[\AURAStreamMetricWidth][c]{1.4} & \makebox[\AURAStreamMetricWidth][c]{0.0} \\

MiniCPM-o~4.5
& \makebox[\AURAStreamMetricWidth][c]{22.1} & \makebox[\AURAStreamMetricWidth][c]{9.8} & \makebox[\AURAStreamMetricWidth][c]{10.5} & \makebox[\AURAStreamMetricWidth][c]{13.3} & \makebox[\AURAStreamMetricWidth][c]{7.6} & \makebox[\AURAStreamMetricWidth][c]{5.0} & \makebox[\AURAStreamMetricWidth][c]{17.1} & \makebox[\AURAStreamMetricWidth][c]{7.5} & \makebox[\AURAStreamMetricWidth][c]{7.3} & \makebox[\AURAStreamMetricWidth][c]{8.1} & \makebox[\AURAStreamMetricWidth][c]{\textbf{5.4}} \\

VST
& \makebox[\AURAStreamMetricWidth][c]{24.0} & \makebox[\AURAStreamMetricWidth][c]{21.2} & \makebox[\AURAStreamMetricWidth][c]{22.1} & \makebox[\AURAStreamMetricWidth][c]{24.8} & \makebox[\AURAStreamMetricWidth][c]{23.2} & \makebox[\AURAStreamMetricWidth][c]{11.6} & \makebox[\AURAStreamMetricWidth][c]{--} & \makebox[\AURAStreamMetricWidth][c]{--} & \makebox[\AURAStreamMetricWidth][c]{--} & \makebox[\AURAStreamMetricWidth][c]{--} & \makebox[\AURAStreamMetricWidth][c]{--} \\

StreamForest
& \makebox[\AURAStreamMetricWidth][c]{17.9} & \makebox[\AURAStreamMetricWidth][c]{14.4} & \makebox[\AURAStreamMetricWidth][c]{19.3} & \makebox[\AURAStreamMetricWidth][c]{13.6} & \makebox[\AURAStreamMetricWidth][c]{14.1} & \makebox[\AURAStreamMetricWidth][c]{11.0} & \makebox[\AURAStreamMetricWidth][c]{--} & \makebox[\AURAStreamMetricWidth][c]{--} & \makebox[\AURAStreamMetricWidth][c]{--} & \makebox[\AURAStreamMetricWidth][c]{--} & \makebox[\AURAStreamMetricWidth][c]{--} \\

ThinkStream
& \makebox[\AURAStreamMetricWidth][c]{8.0} & \makebox[\AURAStreamMetricWidth][c]{7.5} & \makebox[\AURAStreamMetricWidth][c]{12.2} & \makebox[\AURAStreamMetricWidth][c]{7.9} & \makebox[\AURAStreamMetricWidth][c]{5.4} & \makebox[\AURAStreamMetricWidth][c]{4.4} & \makebox[\AURAStreamMetricWidth][c]{1.8} & \makebox[\AURAStreamMetricWidth][c]{1.2} & \makebox[\AURAStreamMetricWidth][c]{2.7} & \makebox[\AURAStreamMetricWidth][c]{0.5} & \makebox[\AURAStreamMetricWidth][c]{0.0} \\

JoyAI-VL-Interaction
& \makebox[\AURAStreamMetricWidth][c]{25.1} & \makebox[\AURAStreamMetricWidth][c]{11.6} & \makebox[\AURAStreamMetricWidth][c]{18.8} & \makebox[\AURAStreamMetricWidth][c]{10.0} & \makebox[\AURAStreamMetricWidth][c]{10.9} & \makebox[\AURAStreamMetricWidth][c]{8.2} & \makebox[\AURAStreamMetricWidth][c]{12.2} & \makebox[\AURAStreamMetricWidth][c]{9.4} & \makebox[\AURAStreamMetricWidth][c]{11.2} & \makebox[\AURAStreamMetricWidth][c]{\textbf{10.2}} & \makebox[\AURAStreamMetricWidth][c]{1.1} \\

MOSS-VL-Realtime
& \makebox[\AURAStreamMetricWidth][c]{23.6} & \makebox[\AURAStreamMetricWidth][c]{8.0} & \makebox[\AURAStreamMetricWidth][c]{8.3} & \makebox[\AURAStreamMetricWidth][c]{9.4} & \makebox[\AURAStreamMetricWidth][c]{8.7} & \makebox[\AURAStreamMetricWidth][c]{4.4} & \makebox[\AURAStreamMetricWidth][c]{--} & \makebox[\AURAStreamMetricWidth][c]{11.8} & \makebox[\AURAStreamMetricWidth][c]{20.1} & \makebox[\AURAStreamMetricWidth][c]{8.8} & \makebox[\AURAStreamMetricWidth][c]{2.2} \\
\midrule
\textbf{AURA 2.0 (Ours)}
& \makebox[\AURAStreamMetricWidth][c]{\textbf{28.5}} & \makebox[\AURAStreamMetricWidth][c]{\textbf{24.5}} & \makebox[\AURAStreamMetricWidth][c]{23.2} & \makebox[\AURAStreamMetricWidth][c]{\textbf{31.5}} & \makebox[\AURAStreamMetricWidth][c]{20.7} & \makebox[\AURAStreamMetricWidth][c]{\textbf{17.0}} & \makebox[\AURAStreamMetricWidth][c]{\textbf{18.5}} & \makebox[\AURAStreamMetricWidth][c]{\textbf{12.8}} & \makebox[\AURAStreamMetricWidth][c]{\textbf{29.7}} & \makebox[\AURAStreamMetricWidth][c]{5.2} & \makebox[\AURAStreamMetricWidth][c]{0.0} \\

\bottomrule
\end{tabular*}%
}

\label{tab:streamarena_results}
\endgroup
\end{table*}

We supplement the published StreamArena results with our own evaluations of three open-source streaming models: JoyAI-VL-Interaction, MOSS-VL-Realtime, and AURA 2.0.
StreamArena consists of 243 videos and evaluates streaming video understanding across four dimensions: 
Real-Time Perception (RTP) measures the ability to answer questions about the current video content in real time, 
Historical Retrospection (HR) evaluates the ability to retrieve and reason about historical visual information across four temporal horizons (L1--L4), 
External Tool Use (Tool) assesses the ability to invoke external tools for tasks requiring additional information or capabilities, 
and Proactive Interaction (Pro) evaluates whether a model can proactively respond when a specified event occurs, while remaining silent before the event.
All results are reported as accuracies (\%), following the official StreamArena evaluation protocol.

As shown in Table~\ref{tab:streamarena_results}, AURA 2.0 achieves 28.5\% in Real-Time Perception and 24.5\% in Historical Retrospection, outperforming JoyAI-VL-Interaction and MOSS-VL-Realtime on both dimensions. Its accuracy on the longest historical horizon (L4) reaches 17.0\%, exceeding AURA and the other listed open-source streaming models. The largest gain over AURA is in Proactive Interaction, where accuracy increases from 3.7\% to 12.8\%, also surpassing JoyAI-VL-Interaction (9.4\%) and MOSS-VL-Realtime (11.8\%). In particular, L1 accuracy rises from 8.9\% to 29.7\%. For External Tool Use, AURA 2.0 achieves 18.5\%, surpassing JoyAI-VL-Interaction (12.2\%) and MiniCPM-o~4.5 (17.1\%). 

\subsubsection{Evaluation on ProactiveVideoQA}

\begin{table*}[!t]
    \centering
    \caption{\textbf{Main results on ProactiveVideoQA.} 
    We report the performance of AURA 2.0 and representative baselines across four categories: WEB, EGO, TV, and VAD.
    The published baseline results, including the human reference, are reproduced from the original ProactiveVideoQA paper.
    Results for JoyAI-VL-Interaction and AURA 2.0 are obtained from our evaluations, while the MOSS-VL-Realtime results are taken from its official technical report.
    All scores are reported using Proactive Area Under Curve (PAUC), where higher values indicate better performance.
    Boldface indicates the best result among open-source streaming models in each metric column.
    }
    \begingroup

    \setlength{\tabcolsep}{4pt}
    \renewcommand{\arraystretch}{1.12}

\sbox0{\AURAComparisonFontReference}
\edef\AURATargetTableWidth{\the\wd0}
\setlength{\tabcolsep}{2pt}
\resizebox{\textwidth}{!}{%
\begin{tabular*}{\AURATargetTableWidth}{@{}@{\extracolsep{\fill}}l*{12}{c}@{}}
        \toprule
        \textbf{Model}
        & \multicolumn{3}{c}{\textbf{WEB}}
        & \multicolumn{3}{c}{\textbf{EGO}}
        & \multicolumn{3}{c}{\textbf{TV}}
        & \multicolumn{3}{c}{\textbf{VAD}}
        \\
        \cmidrule(lr){2-4}
        \cmidrule(lr){5-7}
        \cmidrule(lr){8-10}
        \cmidrule(l){11-13}

        $\omega =$
        & 0.0 & 0.5 & 1.0
        & 0.0 & 0.5 & 1.0
        & 0.0 & 0.5 & 1.0
        & 0.0 & 0.5 & 1.0
        \\
        \midrule

        Human
        & 36.5 & 38.6 & 40.7
        & 35.0 & 38.2 & 41.3
        & 38.1 & 47.0 & 55.9
        & 47.4 & 53.6 & 59.8
        \\

        \midrule
        \multicolumn{13}{@{}l}{\textit{Proprietary Offline Models}}
        \\

        GPT-4.1
        & 44.6 & 51.7 & 58.9
        & 53.6 & 58.8 & 64.0
        & 45.0 & 56.8 & 68.5
        & 40.8 & 46.2 & 51.6
        \\
        
        GPT-4.1-mini
        & 41.2 & 47.8 & 54.5
        & 59.1 & 65.8 & 72.5
        & 48.5 & 59.4 & 70.3
        & 41.4 & 47.7 & 54.0
        \\

        Gemini-1.5-pro
        & 37.1 & 42.1 & 47.0
        & 47.0 & 49.7 & 52.4
        & 41.7 & 52.0 & 62.4
        & 34.3 & 36.1 & 37.9
        \\

        Gemini-2.0-flash
        & 36.1 & 41.0 & 45.9
        & 49.4 & 53.7 & 57.9
        & 40.4 & 49.1 & 57.8
        & 32.9 & 35.8 & 38.6
        \\

        \midrule
        \multicolumn{13}{@{}l}{\textit{Open-source Offline Models}}
        \\

        InternVL-2.5 8B
        & 41.6 & 48.2 & 54.9
        & 52.1 & 57.8 & 63.6
        & 36.5 & 41.5 & 46.6
        & 22.2 & 21.5 & 20.8
        \\
        
        LLaVA-OV 7B
        & 46.6 & 55.0 & 63.4
        & 57.0 & 61.6 & 66.1
        & 38.2 & 45.1 & 51.9
        & 25.3 & 25.6 & 25.9
        \\
        
        LongVA 7B
        & 39.3 & 47.2 & 55.1
        & 34.5 & 37.4 & 40.2
        & 35.3 & 41.5 & 47.6
        & 27.2 & 29.8 & 32.3
        \\
        
        Qwen2.5-VL 7B
        & 45.7 & 52.7 & 59.8
        & 42.8 & 46.5 & 50.3
        & 32.7 & 36.5 & 40.2
        & 27.9 & 29.3 & 30.7
        \\
        
        IXC-2.5 7B
        & 43.0 & 50.3 & 57.6
        & 43.0 & 50.3 & 57.6
        & 41.0 & 48.5 & 56.0
        & 25.4 & 26.8 & 28.3
        \\

        \midrule
        \multicolumn{13}{@{}l}{\textit{Open-source Streaming Models}}
        \\

        MMDuet
        & 37.2 & 38.9 & 40.7
        & 44.0 & 46.0 & 47.9
        & 20.7 & 21.1 & 21.6
        & 26.4 & 27.4 & 28.5
        \\
        
        MMDuet+rm.ass.turns
        & 41.4 & 43.5 & 45.6
        & 49.4 & 52.2 & 55.0
        & 28.2 & 32.6 & 37.1
        & 38.5 & \textbf{42.5} & \textbf{46.5}
        \\
        
        VideoLLM-Online
        & 25.9 & 25.9 & 25.9
        & 25.0 & 25.0 & 25.1
        & 17.8 & 18.3 & 18.8
        & 25.0 & 25.0 & 25.0
        \\

        JoyAI-VL-Interaction
        & 38.7 & 40.7 & 42.7
        & 37.1 & 41.2 & 45.4
        & 33.0 & 34.6 & 36.2
        & 29.6 & 31.3 & 32.9
        \\
        
        MOSS-VL-Realtime
        & -- & 55.4 & --
        & -- & 47.8 & --
        & -- & \textbf{50.2} & --
        & -- & 35.3 & --
        \\
        \midrule
        \textbf{AURA 2.0 (Ours)}
        & \textbf{53.1} & \textbf{58.8} & \textbf{64.5}
        & \textbf{50.1} & \textbf{53.1} & \textbf{56.0}
        & \textbf{43.5} & 49.3 & \textbf{55.0}
        & \textbf{39.1} & 41.8 & 44.4
        \\

        \bottomrule
    \end{tabular*}%
}
    \endgroup

    \label{tab:proactivevideoqa_results}
\end{table*}

Alongside the baselines from the ProactiveVideoQA paper, we report the performance of JoyAI-VL-Interaction, MOSS-VL-Realtime, and AURA 2.0.
The results for MOSS-VL-Realtime are taken from its official technical report, which only provides scores at $\omega=0.5$.
ProactiveVideoQA consists of 1,427 videos across four categories: Web Videos (WEB), Egocentric Videos (EGO), TV Series (TV), and Video Anomaly Detection (VAD).
WEB evaluates general video understanding, while EGO focuses on first-person visual understanding in daily activities.
TV assesses the comprehension of dialogue and social relationships, whereas VAD evaluates the ability to detect and report abnormal events in surveillance videos.
All results are reported using Proactive Area Under Curve (PAUC), which jointly measures response correctness and timeliness.
Specifically, the parameter $\omega$ controls the relative importance of response timing and content correctness.
At $\omega=0$, the metric emphasizes timely responses, while $\omega=1$ evaluates response correctness without considering timing.
The default setting $\omega=0.5$ balances these two aspects.

As shown in Table~\ref{tab:proactivevideoqa_results}, AURA 2.0 demonstrates strong proactive video understanding capabilities across all four categories.
Under the default $\omega=0.5$ setting, it achieves PAUC scores of 58.8, 53.1, 49.3, and 41.8 on WEB, EGO, TV, and VAD, respectively.
It outperforms JoyAI-VL-Interaction (40.7 / 41.2 / 34.6 / 31.3) across all four categories and surpasses MOSS-VL-Realtime (55.4 / 47.8 / 50.2 / 35.3) on WEB, EGO, and VAD.
These results highlight the effectiveness of AURA 2.0 in proactive video understanding, particularly in general web-video comprehension, where it achieves the best WEB score among all listed models at every $\omega$.

\begin{table*}[!t]
    \centering
    \caption{\textbf{Main results on PhoStream.}
    We report the performance of AURA 2.0 and representative baselines across three task categories: Instant, Backward, and Forward.
    Published baseline results are reproduced from the original PhoStream paper, while results for AURA 2.0, JoyAI-VL-Interaction, MOSS-VL-Realtime, and Qwen3.5-9B are obtained from our evaluations.
    Evaluation scores are reported on a 0--100 scale, and Overall denotes the reported aggregate score.
    Boldface indicates the best result among open-source streaming models in each metric column, with lower values preferred for ER and NR and higher values preferred for the other metrics; arrows indicate the preferred direction.
    }
    \label{tab:phostream_results}

    \begingroup

    \setlength{\tabcolsep}{4pt}
    \renewcommand{\arraystretch}{1.12}

    \sbox0{\AURAComparisonFontReference}
    \edef\AURATargetTableWidth{\the\wd0}
    \sbox0{\begin{tabular}{@{}l@{}}Qwen3-Omni-30B-A3B\\Qwen3-VL-30B-A3B\\JoyAI-VL-Interaction\\MOSS-VL-Realtime\\VideoLLM-online-8B\\\textbf{AURA 2.0 (Ours)}\end{tabular}}
    \dimen2=\AURATargetTableWidth\relax
    \advance\dimen2 by -\wd0
    \advance\dimen2 by -56pt
    \divide\dimen2 by 7
    \edef\AURAPhoMetricWidth{\the\dimen2}
    \resizebox{\textwidth}{!}{%
    \begin{tabular*}{\AURATargetTableWidth}{@{}l*{7}{c}@{}}
        \toprule
        \multirow{2}{*}{\textbf{Model}}
        & \multicolumn{3}{c}{\textbf{Evaluation Score} ($\uparrow$)}
        & \makebox[\AURAPhoMetricWidth][c]{\multirow{2}{*}{\textbf{Overall} ($\uparrow$)}} & \multicolumn{3}{c}{\textbf{Forward Task Analysis (\%)}}
        \\
        \cmidrule(lr){2-4}
        \cmidrule(l){6-8}
        & \makebox[\AURAPhoMetricWidth][c]{Instant} & \makebox[\AURAPhoMetricWidth][c]{Backward} & \makebox[\AURAPhoMetricWidth][c]{Forward} & \makebox[\AURAPhoMetricWidth][c]{} & \makebox[\AURAPhoMetricWidth][c]{ER ($\downarrow$)} & \makebox[\AURAPhoMetricWidth][c]{NR ($\downarrow$)} & \makebox[\AURAPhoMetricWidth][c]{PC ($\uparrow$)} \\
        \midrule

        \multicolumn{8}{@{}l}{
            \textit{Proprietary Offline Models}
        } \\

        Gemini 3 Pro
        & \makebox[\AURAPhoMetricWidth][c]{80.83} & \makebox[\AURAPhoMetricWidth][c]{82.19} & \makebox[\AURAPhoMetricWidth][c]{16.40} & \makebox[\AURAPhoMetricWidth][c]{48.70} & \makebox[\AURAPhoMetricWidth][c]{79.12} & \makebox[\AURAPhoMetricWidth][c]{0.11} & \makebox[\AURAPhoMetricWidth][c]{20.77} \\

        Doubao-Seed-1.6
        & \makebox[\AURAPhoMetricWidth][c]{71.28} & \makebox[\AURAPhoMetricWidth][c]{62.94} & \makebox[\AURAPhoMetricWidth][c]{44.26} & \makebox[\AURAPhoMetricWidth][c]{56.01} & \makebox[\AURAPhoMetricWidth][c]{29.76} & \makebox[\AURAPhoMetricWidth][c]{13.38} & \makebox[\AURAPhoMetricWidth][c]{56.85} \\

        Doubao-Seed-1.8
        & \makebox[\AURAPhoMetricWidth][c]{80.45} & \makebox[\AURAPhoMetricWidth][c]{77.31} & \makebox[\AURAPhoMetricWidth][c]{33.38} & \makebox[\AURAPhoMetricWidth][c]{56.15} & \makebox[\AURAPhoMetricWidth][c]{56.46} & \makebox[\AURAPhoMetricWidth][c]{2.36} & \makebox[\AURAPhoMetricWidth][c]{41.18} \\

        \midrule
        \multicolumn{8}{@{}l}{
            \textit{Open-source Offline Models}
        } \\

        Qwen2.5-Omni-7B
        & \makebox[\AURAPhoMetricWidth][c]{67.71} & \makebox[\AURAPhoMetricWidth][c]{65.20} & \makebox[\AURAPhoMetricWidth][c]{1.81} & \makebox[\AURAPhoMetricWidth][c]{34.06} & \makebox[\AURAPhoMetricWidth][c]{42.65} & \makebox[\AURAPhoMetricWidth][c]{43.50} & \makebox[\AURAPhoMetricWidth][c]{13.85} \\

        Qwen3-VL-8B
        & \makebox[\AURAPhoMetricWidth][c]{75.22} & \makebox[\AURAPhoMetricWidth][c]{71.50} & \makebox[\AURAPhoMetricWidth][c]{7.18} & \makebox[\AURAPhoMetricWidth][c]{40.25} & \makebox[\AURAPhoMetricWidth][c]{85.44} & \makebox[\AURAPhoMetricWidth][c]{3.50} & \makebox[\AURAPhoMetricWidth][c]{11.06} \\

        Qwen3.5-9B
        & \makebox[\AURAPhoMetricWidth][c]{72.13} & \makebox[\AURAPhoMetricWidth][c]{62.92} & \makebox[\AURAPhoMetricWidth][c]{11.18} & \makebox[\AURAPhoMetricWidth][c]{39.63} & \makebox[\AURAPhoMetricWidth][c]{66.52} & \makebox[\AURAPhoMetricWidth][c]{17.70} & \makebox[\AURAPhoMetricWidth][c]{15.77} \\

        Qwen3-VL-30B-A3B
        & \makebox[\AURAPhoMetricWidth][c]{73.38} & \makebox[\AURAPhoMetricWidth][c]{69.46} & \makebox[\AURAPhoMetricWidth][c]{5.25} & \makebox[\AURAPhoMetricWidth][c]{38.33} & \makebox[\AURAPhoMetricWidth][c]{91.33} & \makebox[\AURAPhoMetricWidth][c]{1.18} & \makebox[\AURAPhoMetricWidth][c]{7.49} \\

        Qwen3-Omni-30B-A3B
        & \makebox[\AURAPhoMetricWidth][c]{77.18} & \makebox[\AURAPhoMetricWidth][c]{77.24} & \makebox[\AURAPhoMetricWidth][c]{1.26} & \makebox[\AURAPhoMetricWidth][c]{39.02} & \makebox[\AURAPhoMetricWidth][c]{97.89} & \makebox[\AURAPhoMetricWidth][c]{0.07} & \makebox[\AURAPhoMetricWidth][c]{2.03} \\

        \midrule
        \multicolumn{8}{@{}l}{
            \textit{Open-source Streaming Models}
        } \\

        Dispider
        & \makebox[\AURAPhoMetricWidth][c]{44.24} & \makebox[\AURAPhoMetricWidth][c]{42.90} & \makebox[\AURAPhoMetricWidth][c]{3.53} & \makebox[\AURAPhoMetricWidth][c]{23.50} & \makebox[\AURAPhoMetricWidth][c]{68.52} & \makebox[\AURAPhoMetricWidth][c]{21.20} & \makebox[\AURAPhoMetricWidth][c]{10.28} \\

        VideoLLM-online-8B
        & \makebox[\AURAPhoMetricWidth][c]{24.88} & \makebox[\AURAPhoMetricWidth][c]{24.72} & \makebox[\AURAPhoMetricWidth][c]{0.00} & \makebox[\AURAPhoMetricWidth][c]{12.34} & \makebox[\AURAPhoMetricWidth][c]{99.54} & \makebox[\AURAPhoMetricWidth][c]{\textbf{0.43}} & \makebox[\AURAPhoMetricWidth][c]{0.04} \\

        MMDuet2
        & \makebox[\AURAPhoMetricWidth][c]{8.76} & \makebox[\AURAPhoMetricWidth][c]{8.30} & \makebox[\AURAPhoMetricWidth][c]{1.39} & \makebox[\AURAPhoMetricWidth][c]{4.96} & \makebox[\AURAPhoMetricWidth][c]{28.55} & \makebox[\AURAPhoMetricWidth][c]{59.21} & \makebox[\AURAPhoMetricWidth][c]{12.24} \\

        JoyAI-VL-Interaction
        & \makebox[\AURAPhoMetricWidth][c]{68.35} & \makebox[\AURAPhoMetricWidth][c]{59.48} & \makebox[\AURAPhoMetricWidth][c]{2.40} & \makebox[\AURAPhoMetricWidth][c]{33.40} & \makebox[\AURAPhoMetricWidth][c]{94.93} & \makebox[\AURAPhoMetricWidth][c]{1.32} & \makebox[\AURAPhoMetricWidth][c]{3.75} \\

        MOSS-VL-Realtime
        & \makebox[\AURAPhoMetricWidth][c]{72.69} & \makebox[\AURAPhoMetricWidth][c]{65.54} & \makebox[\AURAPhoMetricWidth][c]{18.27} & \makebox[\AURAPhoMetricWidth][c]{43.88} & \makebox[\AURAPhoMetricWidth][c]{41.51} & \makebox[\AURAPhoMetricWidth][c]{28.87} & \makebox[\AURAPhoMetricWidth][c]{29.62} \\
        \midrule
        \textbf{AURA 2.0 (Ours)}
        & \makebox[\AURAPhoMetricWidth][c]{\textbf{77.89}} & \makebox[\AURAPhoMetricWidth][c]{\textbf{71.48}} & \makebox[\AURAPhoMetricWidth][c]{\textbf{56.22}} & \makebox[\AURAPhoMetricWidth][c]{\textbf{65.69}} & \makebox[\AURAPhoMetricWidth][c]{\textbf{1.30}} & \makebox[\AURAPhoMetricWidth][c]{23.90} & \makebox[\AURAPhoMetricWidth][c]{\textbf{74.80}} \\

        \bottomrule
    \end{tabular*}%
    }
    \endgroup
\end{table*}

\subsubsection{Evaluation on PhoStream}

Beyond the models covered in the PhoStream paper, we further evaluate Qwen3.5-9B, JoyAI-VL-Interaction, MOSS-VL-Realtime, and AURA 2.0.
PhoStream is a mobile-centric streaming video understanding benchmark containing 578 videos and 5,572 open-ended questions across diverse real-world scenarios.
It evaluates three types of temporal understanding: Instant, Backward, and Forward.
Instant measures the ability to answer questions based on the current video content, while Backward evaluates the retrieval and reasoning of previously observed information.
Forward assesses whether a model can continuously monitor the video stream and respond only when sufficient evidence becomes available in the future.
The Overall score is computed as the question-weighted average across the three task categories.
For Forward tasks, we further report Early Response (ER), No Response (NR), and Partly Correct (PC).

Table~\ref{tab:phostream_results} shows that AURA 2.0 achieves the highest reported Overall score of 65.69, exceeding the strongest proprietary baseline, Doubao-Seed-1.8, by 9.54 points.
It leads the open-source streaming models on all three tasks, scoring 77.89 on Instant, 71.48 on Backward, and 56.22 on Forward.
Its largest advantage lies in Forward reasoning, where it exceeds MOSS-VL-Realtime by 37.95 points.
This improvement is accompanied by the lowest Early Response rate among all listed models, at 1.30\%, compared with 41.51\% for MOSS-VL-Realtime, 94.93\% for JoyAI-VL-Interaction, and 66.52\% for Qwen3.5-9B.
AURA 2.0 also achieves the highest Partly Correct rate of 74.80\%, indicating that a larger proportion of responses fall within the valid response window; this category reflects response timing rather than guaranteed semantic correctness.
Its No Response rate remains 23.90\%, highlighting missed responses as a remaining limitation.

\begin{figure}[!t]
\centering
\includegraphics[width=\linewidth]{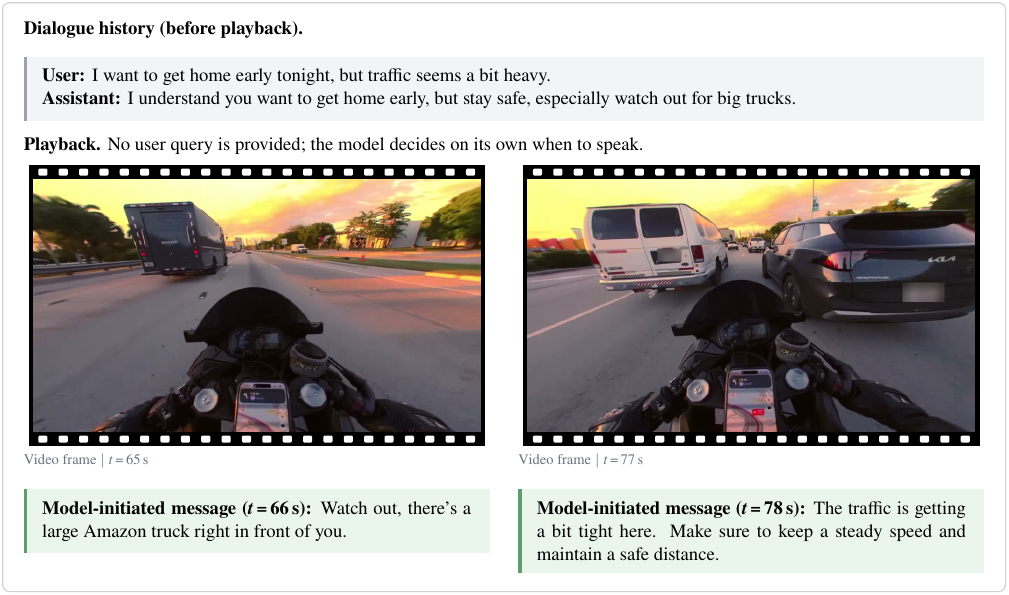}
\caption{\textbf{MIC example in the Visual-Dialogue Context setting.} Before playback, the user mentions wanting to get home early despite heavy traffic, and the assistant advises caution around big trucks. Without any further user query, AURA 2.0 initiates two messages: at $t = 66$\,s it warns about a large Amazon truck ahead, and at $t = 78$\,s it advises keeping a steady speed and a safe distance as traffic tightens. Both messages build on the user's expressed concern and the assistant's earlier advice. Each frame is shown one second before the corresponding message onset.}
\label{fig:mic_case}
\end{figure}

\subsection{Evaluation of Model-Initiated Conversation}\label{sec:MIC_results}

We evaluate MIC on MIC-Bench, an internal benchmark. It contains 25 videos evaluated under two settings, yielding 50 test instances. The \textit{Visual-Dialogue Context} setting tests whether the model can identify opportunities to build on a prior dialogue, while the \textit{Visual-Only Context} setting tests whether it can identify topics worth discussing from visual observations alone. No new user queries are provided during playback in either setting.

We report \textit{trigger rate}, the percentage of evaluation units (annotated windows or full videos) containing at least one nonempty model-initiated message, and \textit{response quality}, which measures the value of those messages. To reflect interruption cost, we also report the number of messages per minute over the entire playback, taking the median across videos. An MLLM judge (\texttt{gpt-4o}) receives the conversational context (when available), the model-initiated message, and two video frames in chronological order, sampled one second before and at message onset. Two annotators reviewed all judge outputs and found no unreasonable judgments.

\noindent\textbf{Visual-Dialogue Context.} The model receives a short prior conversation expressing a user need, interest, or concern. Across all test instances, the dialogue history contains no outstanding user requests or monitoring instructions. Each video has one human-annotated trigger window marking when a visual cue offers an opportunity to build on this context, and the trigger rate is computed over these 25 windows. All messages within a window are jointly scored from 0 to 5 on how specifically and usefully they address the user's previously expressed need, interest, or concern. Quality is averaged over triggered windows; messages outside annotated windows are excluded.

\noindent\textbf{Visual-Only Context.} This setting provides no dialogue history, and evaluation covers the entire video. The judge additionally receives a scene summary and human-verified visual details. Each message is rated \textit{Good} (5), \textit{Neutral} (2.5), or \textit{Bad} (0). Good messages are accurate and provide specific information, actionable help, or meaningful interaction grounded in the scene; Neutral messages offer limited value or contain minor flaws; Bad messages contain a core factual error, valueless repetition, or harmful content. Quality is averaged across messages within each video, then equally across videos; a silent video receives a score of 0.

\begin{table}[!b]
\centering
\small
\caption{\textbf{MIC results on internal MIC-Bench (25 videos).} Msgs/min is the median across videos over the entire playback. The two settings differ in evaluation interval and scoring procedure, so their trigger rates and quality scores are not directly comparable.}
\label{tab:mic_eval}
\begin{tabular}{@{}llccc@{}}
\toprule
\textbf{Setting} & \textbf{Evaluated over} & \textbf{Trigger rate (\%)} & \textbf{Quality (0--5)} & \textbf{Msgs/min} \\
\midrule
Visual-Dialogue Context & Annotated windows & 80.0 (20/25) & 4.10 & 3.0 \\
Visual-Only Context & Entire videos & 96.0 (24/25) & 3.62 & 3.5 \\
\bottomrule
\end{tabular}
\end{table}

Table~\ref{tab:mic_eval} summarizes the MIC performance of AURA 2.0. With dialogue context, the model triggers in 80.0\% of annotated windows (20/25) and achieves a mean quality of 4.10/5, indicating well-targeted contributions, although some relevant opportunities remain unaddressed.  Figure~\ref{fig:mic_case} shows an example in which both model-initiated messages build on the prior dialogue. With visual input only, the trigger rate rises to 96.0\% (24/25), but quality drops to 3.62/5: of the 173 model-initiated messages, 93 are rated Good, 65 Neutral, and 15 Bad. Without an expressed user need or interest to guide topic selection, the model may find it harder to distinguish observations worth sharing from those of limited value. The model speaks 3.0 and 3.5 times per minute in the two settings, respectively.

\subsection{Evaluation of Stream-Aware Tool Calling}
\label{sec:eval_satc}

We evaluate SATC on an internal benchmark containing 167 questions across 55 video clips, covering transportation, landmarks, plants, and other everyday subjects. We compare AURA 2.0 with tools enabled and disabled, using an LLM judge to assess final answers against reference answers. Enabling tools improves accuracy from 35.9\% to 48.5\%, a gain of 12.6 percentage points. The model invokes tools for 41.3\% of questions, making 73 calls in total. These results demonstrate the value of external information for answering questions grounded in the video stream.

\begin{figure}[!t] \centering %
\begin{tikzpicture}
  \node[anchor=south west,inner sep=0] (img) at (0,0) {\includegraphics[width=\linewidth]{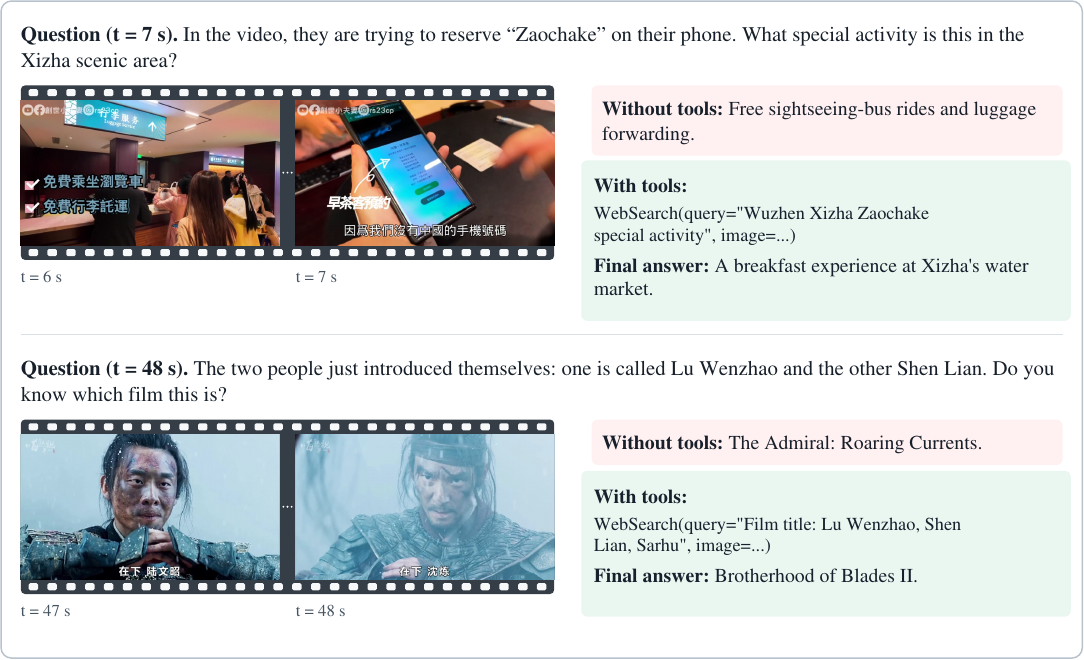}};
  \begin{scope}[x={(img.south east)},y={(img.north west)}]
  \fill[white] (0.0152,0.5600) rectangle (0.3600,0.5960);
  \fill[white] (0.0152,0.0540) rectangle (0.3600,0.0900);
  \fill[fill={rgb,255:red,200;green,45;blue,55}] (0.0188,0.8167) rectangle (0.1176,0.8508);
  \node[text=white,font=\sffamily\bfseries\fontsize{6.5}{7}\selectfont,inner sep=0] at (0.0682,0.8337) {t\,=\,6\,s};
  \fill[fill={rgb,255:red,200;green,45;blue,55}] (0.2728,0.8167) rectangle (0.3717,0.8508);
  \node[text=white,font=\sffamily\bfseries\fontsize{6.5}{7}\selectfont,inner sep=0] at (0.3222,0.8337) {t\,=\,7\,s};
  \fill[fill={rgb,255:red,200;green,45;blue,55}] (0.0188,0.3083) rectangle (0.1176,0.3425);
  \node[text=white,font=\sffamily\bfseries\fontsize{6.5}{7}\selectfont,inner sep=0] at (0.0682,0.3254) {t\,=\,47\,s};
  \fill[fill={rgb,255:red,200;green,45;blue,55}] (0.2728,0.3083) rectangle (0.3717,0.3425);
  \node[text=white,font=\sffamily\bfseries\fontsize{6.5}{7}\selectfont,inner sep=0] at (0.3222,0.3254) {t\,=\,48\,s};
  \end{scope}
\end{tikzpicture}%
 \caption{\textbf{SATC examples from the internal benchmark.} Top: \texttt{WebSearch} identifies the ``Zaochake'' reservation as a breakfast experience at the water market. Bottom: \texttt{WebSearch} uses character names from the subtitles to identify \textit{Brotherhood of Blades II}. Both examples correct answers generated without tools.} \label{fig:satc_internal} \end{figure}

Figure~\ref{fig:satc_internal} illustrates how visual cues guide external retrieval through \texttt{WebSearch}. In the upper example, the model without tools answers a question about a ``Zaochake'' reservation using information about earlier transportation services. \texttt{WebSearch} instead identifies it as a breakfast experience at Xizha's water market. In the lower example, \texttt{WebSearch} uses the character names Lu Wenzhao and Shen Lian, visible in the subtitles, to identify \textit{Brotherhood of Blades II}, correcting the model's answer of \textit{The Admiral: Roaring Currents}. In both cases, retrieval helps resolve an incorrect interpretation using cues from the current visual context.

\begin{figure}[!t] \centering \includegraphics[width=\linewidth]{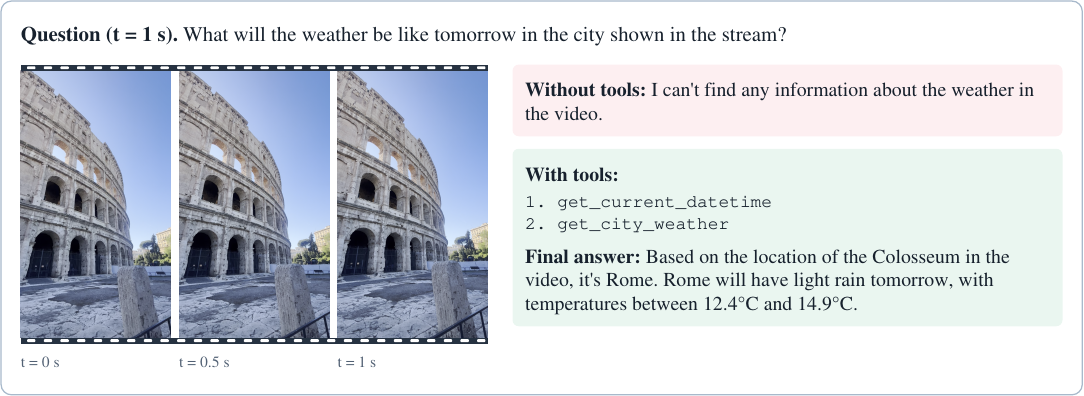} \caption{\textbf{Real-world weather query at $t=1$\,s.} AURA 2.0 identifies Rome from the Colosseum and calls \texttt{get\_current\_datetime} and \texttt{get\_city\_weather} to retrieve the forecast. Without tools, the model reports that weather information is unavailable in the video; with tools, it provides the forecast for the visually identified city.} \label{fig:satc_rome_weather} \end{figure}

We further present a real-world weather query in Figure~\ref{fig:satc_rome_weather}. At $t=1$\,s, the user asks, ``What will the weather be like tomorrow in the city shown in the stream?'' Without tools, the model reports that weather information is unavailable in the video. With tools enabled, AURA 2.0 identifies Rome from the Colosseum, calls \texttt{get\_current\_datetime} and \texttt{get\_city\_weather}, and reports light rain with temperatures ranging from $12.4\,^{\circ}\mathrm{C}$ to $14.9\,^{\circ}\mathrm{C}$. This example illustrates how SATC combines visual recognition with time-sensitive external information.

\subsection{Evaluation of Interactive Visual Grounding}
\label{sec:eval_ivg}

We evaluate IVG's bounding-box outputs on a diagnostic set of 50 COCO~\citep{lin2014coco} images, comparing AURA 2.0 with its backbone, Qwen3.5-9B~\citep{qwen2026qwen35}.
The queries cover both present and absent targets, allowing us to assess localization and abstention.
We report mean intersection over union (mIoU) for present targets and accuracy across all queries.

\begin{table}[!b]
\centering
\small
\setlength{\tabcolsep}{5pt}
\renewcommand{\arraystretch}{1.18}
\caption{\textbf{IVG results on a 50-image diagnostic set.} mIoU is computed on present targets, with missing boxes assigned zero. Positive hits require IoU $\geq 0.5$; correct abstentions require no box for an absent target.}
\label{tab:ivg_eval}
\begin{tabular*}{\linewidth}{@{\extracolsep{\fill}}lccccc@{}}
\toprule
\textbf{Model} & \textbf{mIoU} & \shortstack{\textbf{Accuracy}\\\textbf{(\%)}} & \shortstack{\textbf{Positive queries}\\\textbf{with a box}} & \shortstack{\textbf{Positive}\\\textbf{hits}} & \shortstack{\textbf{Correct}\\\textbf{abstentions}} \\
\midrule
Qwen3.5-9B & 0.635 & 76.0 & 26/37 & 25/37 & \textbf{13/13} \\
\textbf{AURA 2.0 (Ours)} & \textbf{0.906} & \textbf{94.0} & \textbf{37/37} & \textbf{36/37} & 11/13 \\
\bottomrule
\end{tabular*}
\end{table}

\begin{figure}[!t]
\centering
\includegraphics[width=\linewidth]{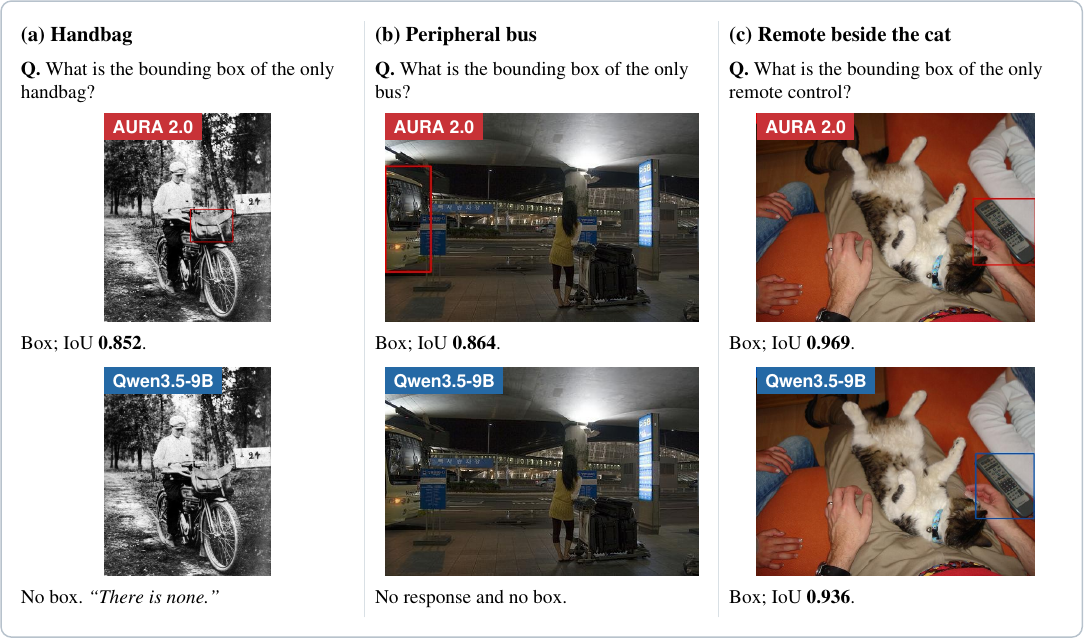}
\caption{\textbf{IVG examples with less salient and peripheral targets.} AURA 2.0 (top, red boxes) localizes both the handbag and the partially visible bus. Qwen3.5-9B (bottom, blue boxes) incorrectly declares the handbag absent and produces no response for the bus. Both models accurately localize the remote control despite the salient cat.}
\label{fig:ivg_cases}
\end{figure}

As shown in Table~\ref{tab:ivg_eval}, AURA 2.0 achieves an mIoU of 0.906 and an accuracy of 94.0\%, versus 0.635 and 76.0\% for Qwen3.5-9B. It returns boxes for all 37 positive queries, versus the backbone's 26. On the 26 queries where both models return boxes, mean IoUs are similar (0.906 vs.\ 0.903, respectively), indicating that the overall improvement comes primarily from fewer missing outputs. For absent targets, AURA 2.0 correctly abstains on 11 of 13 queries, versus all 13 for the backbone.

Figure~\ref{fig:ivg_cases} illustrates this improvement on less salient and partially visible targets. AURA 2.0 localizes the handbag and bus missed by the backbone, while both models identify the remote control beside the cat. These examples support the quantitative finding that more consistent bounding-box output improves coverage of the requested targets.

\subsection{Inference Efficiency}
\label{sec:exp_efficiency}

All measurements are run on a server with two accelerators of about 140\,GB memory each. The language model occupies one accelerator; speech recognition, speech synthesis, and codec decoding share the other. We stream videos from OmniInteract~\citep{lu-etal-2026-omniinteract} in real time and inject each spoken query as audio at its annotated time. With multiple concurrent sessions, each session starts at a random offset within the first 10\,s. Each concurrency level is run three times; we pool the runs and report medians over turns in which the model responds with speech.

All times except Time Per Output Token (TPOT) are measured at the client, starting from the moment the client submits a turn (the end of the spoken query).
Time to First Token (TTFT) is the time until the first response text arrives, including speech recognition.
Time To First Packet (TTFP) is the time until the first reply audio packet arrives.
Real-Time Factor (RTF) is the time until the last audio packet arrives, divided by the duration of the received reply audio; values below 1 indicate that speech is delivered faster than it is played.
TPOT covers only the language-model stage and is reported by the serving engine.
\begin{table}[!b]
\centering
\caption{\textbf{Latency under concurrent sessions (median over pooled turns).} TTFT, TPOT, and TTFP are in milliseconds; RTF is unitless. Lower is better for all metrics.}
\label{tab:latency_conc}
\begin{tabular}{lcccc}
\toprule
\textbf{Concurrent sessions }& \textbf{TTFT} & \textbf{TPOT} & \textbf{TTFP} & \textbf{RTF} \\
\midrule
1 & 157 & 7.6 & 344 & 0.135 \\
8 & 184 & 12.8 & 449 & 0.161 \\
16 & 234 & 21.1 & 601 & 0.203 \\
\bottomrule
\end{tabular}
\end{table}
With a single session, the first text arrives after 157\,ms and the first audio packet after 344\,ms. Even with 16 concurrent sessions, the first audio packet arrives within about 0.6\,s, and the median RTF remains well below 1.

\begin{figure}[t]
\centering
\includegraphics[width=\linewidth]{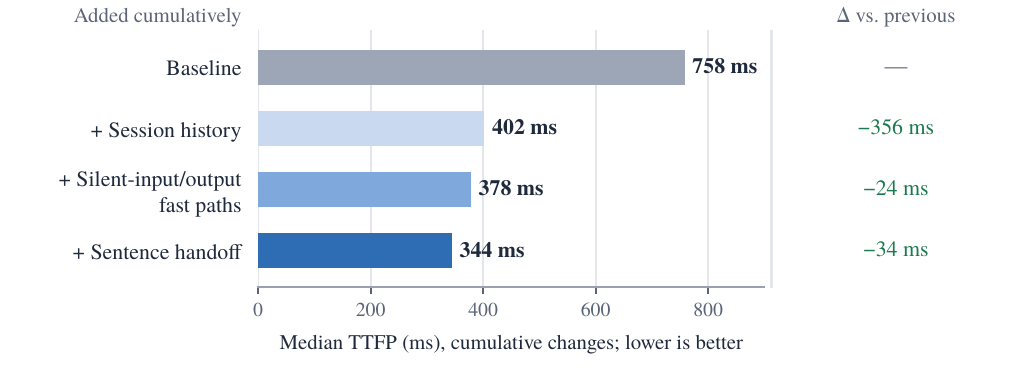}
\caption{\textbf{Cumulative effect of the inference optimizations on median TTFP with a single session.} Each row adds one optimization to the preceding configuration, and the full configuration reduces median TTFP by 54.6\%. The right column reports the change relative to the previous row.}
\label{fig:latency_ablation}
\end{figure}

Figure~\ref{fig:latency_ablation} shows the median TTFP with one session as the three optimizations in Section~\ref{sec:latency_opt} are added cumulatively. Together, they reduce the median TTFP from 758\,ms to 344\,ms (54.6\%), with session history accounting for most of the gain (about 86\%).

\section{Conclusion and Limitations}
\label{sec:conclusion}

We present AURA 2.0, a streaming VideoLLM that extends query-driven interaction with model-initiated conversation. Its core capability, MIC, enables the model to identify topics worth discussing and initiate conversations based on evolving visual observations and prior dialogue. SATC lets the model act on the visual and conversational context through external tools, and IVG supports visually grounded interaction. A dedicated streaming interaction dataset and three-stage training scheme integrate these capabilities with existing streaming QA behaviors. LHIM preserves historical events as compact textual memories, while serving optimizations reduce response latency. Evaluations on four public streaming benchmarks demonstrate strong performance in video understanding and response timing. Targeted evaluations show useful MIC openings, improved answer accuracy with tool access, and fewer missing bounding-box outputs for requested targets. Together, these results demonstrate progress toward streaming assistants that both respond to user requests and identify opportunities to initiate conversations.

The usefulness of MIC is inherently subjective and depends on individual preferences for topic, timing, and speaking frequency. Future work should incorporate user preferences and broader interaction context to better determine when and how to engage. Our current system uses a cascaded ASR--VideoLLM--TTS pipeline, which we plan to extend to an omni-modal, full-duplex model. For example, during piano practice, such a model could listen to the user's performance and observe their hand movements while providing spoken guidance. It could adjust its ongoing feedback upon hearing a wrong note and identifying the corresponding fingering error. This simultaneous listening, viewing, and speaking is not supported by our current system.

\clearpage
\bibliographystyle{unsrt}
\bibliography{references_formated}

@article{thinkingmachines2026interactionmodels,
  author={{Thinking Machines Lab}},
  title={Interaction Models: A Scalable Approach to Human-AI Collaboration},
  journal={Thinking Machines Lab: Connectionism},
  year={2026},
  month=may,
  note={https://thinkingmachines.ai/blog/interaction-models/},
  doi={10.64434/tml.20260511}
}

@article{yao2026joyai,
  title={{JoyAI-VL-Interaction}: Real-Time Vision-Language Interaction Intelligence},
  author={Yao, Dingyu and Zhou, Junhao and Yang, Chenxu and Qin, Chuanyu and Zeng, Xiangyu and Li, Yifei and Hou, Haowen and Liang, Zheming and Wang, Congcong and Tuo, Kaiwen and Zhang, Jun and Zhu, Yuhan and Cao, Yuhang and Ye, Shenglong and Xie, Shuai and Gu, Shuhuan and Huang, Haoyang and Si, Qingyi and Duan, Nan and Wang, Jiaqi},
  journal={arXiv preprint arXiv:2606.14777},
  year={2026}
}

@article{zheng2026emgarde,
  title={{Em-Garde}: A Propose-Match Framework for Proactive Streaming Video Understanding},
  author={Zheng, Yikai and Ding, Xin and Yang, Yifan and Jiang, Shiqi and Wu, Hao and Zhang, Qianxi and Wang, Weijun and Cao, Ting and Liu, Yunxin},
  journal={arXiv preprint arXiv:2603.19054},
  year={2026}
}

@article{yang2026livestarpro,
  title={{LiveStarPro}: Proactive Streaming Video Understanding with Hierarchical Memory for Long-Horizon Streams},
  author={Yang, Zhenyu and Zhang, Kairui and Wang, Bing and Qian, Shengsheng and Xu, Changsheng},
  journal={arXiv preprint arXiv:2606.17798},
  year={2026}
}

@article{yao2026harnessing,
  title={Harnessing Streaming Video in the Wild},
  author={Yao, Dingyu and Gu, Shuhuan and Si, Qingyi and Zhou, Junhao and Yang, Chenxu and Qin, Chuanyu and Gu, Naibin and Lin, Zheng and Wang, Weiping and Duan, Nan and Wang, Jiaqi},
  journal={arXiv preprint arXiv:2606.08615},
  year={2026}
}

@article{zeng2026onestreamer,
  title={{OneStreamer}: Unifying Perception, Memory, and Proactive Response in Streaming Video Interaction},
  author={Zeng, Xiangyu and Yang, Yuandong and Zhang, Zhiqiu and Zhu, Yuhan and Li, Xinhao and Si, Qingyi and Yao, Dingyu and Ma, Changlian and Chen, Haoran and Chen, Xinyu and Shi, Yansong and Zhou, Junhao and Li, Yifei and Zhang, Jun and Qin, Chuanyu and Yang, Chenxu and Yu, Xinlei and Ouyang, Kun and Shao, Yuchen and Wei, Qianshan and Zhou, Changhai and Gao, Jun and Wang, Jiaqi and Wang, Limin},
  journal={arXiv preprint arXiv:2610.01762},
  year={2026}
}

@article{wang2025streambridge,
  title={{StreamBridge}: Turning Your Offline Video Large Language Model into a Proactive Streaming Assistant},
  author={Wang, Haibo and Feng, Bo and Lai, Zhengfeng and Xu, Mingze and Li, Shiyu and Ge, Weifeng and Dehghan, Afshin and Cao, Meng and Huang, Ping},
  journal={arXiv preprint arXiv:2505.05467},
  year={2025}
}

@article{lu2026aura,
  title={{AURA}: Always-On Understanding and Real-Time Assistance via Video Streams},
  author={Lu, Xudong and Bo, Yang and Chen, Jinpeng and Li, Shuhan and Guo, Xintong and Guan, Huankang and Liu, Fang and Xu, Dunyuan and Sun, Peiwen and Sun, Heyang and Liu, Rui and Li, Hongsheng},
  journal={arXiv preprint arXiv:2604.04184},
  year={2026}
}

@inproceedings{niu2025ovo,
  title={{OVO-Bench}: How far is your video-llms from real-world online video understanding?},
  author={Niu, Junbo and Li, Yifei and Miao, Ziyang and Ge, Chunjiang and Zhou, Yuanhang and He, Qihao and Dong, Xiaoyi and Duan, Haodong and Ding, Shuangrui and Qian, Rui and others},
  booktitle={Proc. CVPR},
  pages={18902--18913},
  year={2025}
}

@article{lin2024streamingbench,
  title={{StreamingBench}: Assessing the gap for mllms to achieve streaming video understanding},
  author={Lin, Junming and Fang, Zheng and Chen, Chi and Wan, Zihao and Luo, Fuwen and Li, Peng and Liu, Yang and Sun, Maosong},
  journal={arXiv preprint arXiv:2411.03628},
  year={2024}
}

@inproceedings{wang2025omnimmi,
  title={{OmniMMI}: A comprehensive multi-modal interaction benchmark in streaming video contexts},
  author={Wang, Yuxuan and Wang, Yueqian and Chen, Bo and Wu, Tong and Zhao, Dongyan and Zheng, Zilong},
  booktitle={Proc. CVPR},
  pages={18925--18935},
  year={2025}
}

@article{bai2025qwen3,
  title={{Qwen3-VL} Technical Report},
  author={Bai, Shuai and Cai, Yuxuan and Chen, Ruizhe and Chen, Keqin and Chen, Xionghui and Cheng, Zesen and Deng, Lianghao and Ding, Wei and Gao, Chang and Ge, Chunjiang and others},
  journal={arXiv preprint arXiv:2511.21631},
  year={2025}
}

@article{wang2025internvl3,
  title={{InternVL3.5}: Advancing open-source multimodal models in versatility, reasoning, and efficiency},
  author={Wang, Weiyun and Gao, Zhangwei and Gu, Lixin and Pu, Hengjun and Cui, Long and Wei, Xingguang and Liu, Zhaoyang and Jing, Linglin and Ye, Shenglong and Shao, Jie and others},
  journal={arXiv preprint arXiv:2508.18265},
  year={2025}
}

@inproceedings{chen2024videollm,
  title={{VideoLLM-Online}: Online video large language model for streaming video},
  author={Chen, Joya and Lv, Zhaoyang and Wu, Shiwei and Lin, Kevin Qinghong and Song, Chenan and Gao, Difei and Liu, Jia-Wei and Gao, Ziteng and Mao, Dongxing and Shou, Mike Zheng},
  booktitle={Proc. CVPR},
  pages={18407--18418},
  year={2024}
}

@article{zhang2024flash,
  title={{Flash-VStream}: Memory-based real-time understanding for long video streams},
  author={Zhang, Haoji and Wang, Yiqin and Tang, Yansong and Liu, Yong and Feng, Jiashi and Dai, Jifeng and Jin, Xiaojie},
  journal={arXiv preprint arXiv:2406.08085},
  year={2024}
}

@article{wang2025proactivevideoqa,
  title={{ProactiveVideoQA}: A comprehensive benchmark evaluating proactive interactions in video large language models},
  author={Wang, Yueqian and Meng, Xiaojun and Wang, Yifan and Zhang, Huishuai and Zhao, Dongyan},
  journal={arXiv preprint arXiv:2507.09313},
  year={2025}
}

@article{li2026ipibench,
  title={{IPIBench}: Evaluating Interactive Proactive Intelligence of {MLLMs} under Continuous Streams},
  author={Li, Jinzhao and Chen, Yinuo and Song, Wenxuan and Lei, Yijia and Zhang, Yichi and Yan, Honglei and Pan, Panwang and Liu, Miao},
  journal={arXiv preprint arXiv:2605.27074},
  year={2026}
}

@article{lu2026phostream,
  title={{PhoStream}: Benchmarking Real-World Streaming for Omnimodal Assistants in Mobile Scenarios},
  author={Lu, Xudong and Guan, Huankang and Bo, Yang and Chen, Jinpeng and Guo, Xintong and Li, Shuhan and Liu, Fang and Sun, Peiwen and Li, Xueying and Zhang, Wei and Yang, Xue and Liu, Rui and Li, Hongsheng},
  journal={arXiv preprint arXiv:2601.22575},
  year={2026}
}

@inproceedings{qian2025dispider,
  title={Dispider: Enabling video llms with active real-time interaction via disentangled perception, decision, and reaction},
  author={Qian, Rui and Ding, Shuangrui and Dong, Xiaoyi and Zhang, Pan and Zang, Yuhang and Cao, Yuhang and Lin, Dahua and Wang, Jiaqi},
  booktitle={Proc. CVPR},
  pages={24045--24055},
  year={2025}
}

@article{wang2025mmduet2,
  title={{MMDuet2}: Enhancing Proactive Interaction of Video MLLMs with Multi-Turn Reinforcement Learning},
  author={Wang, Yueqian and Liu, Songxiang and Wang, Disong and Xu, Nuo and Wan, Guanglu and Zhang, Huishuai and Zhao, Dongyan},
  journal={arXiv preprint arXiv:2512.06810},
  year={2025}
}

@inproceedings{chen2025livecc,
  title={{LiveCC}: Learning video llm with streaming speech transcription at scale},
  author={Chen, Joya and Zeng, Ziyun and Lin, Yiqi and Li, Wei and Ma, Zejun and Shou, Mike Zheng},
  booktitle={Proc. CVPR},
  pages={29083--29095},
  year={2025}
}

@inproceedings{yao2025timechat,
  title={{TimeChat-Online}: 80\% visual tokens are naturally redundant in streaming videos},
  author={Yao, Linli and Li, Yicheng and Wei, Yuancheng and Li, Lei and Ren, Shuhuai and Liu, Yuanxin and Ouyang, Kun and Wang, Lean and Li, Shicheng and Li, Sida and others},
  booktitle={Proc. ACM MM},
  pages={10807--10816},
  year={2025}
}

@article{yang2025kwai,
  title={{Kwai Keye-VL 1.5} Technical Report},
  author={Yang, Biao and Wen, Bin and Ding, Boyang and Liu, Changyi and Chu, Chenglong and Song, Chengru and Rao, Chongling and Yi, Chuan and Li, Da and Zang, Dunju and others},
  journal={arXiv preprint arXiv:2509.01563},
  year={2025}
}

@inproceedings{fu2025vispeak,
  title={{ViSpeak}: Visual instruction feedback in streaming videos},
  author={Fu, Shenghao and Yang, Qize and Li, Yuan-Ming and Peng, Yi-Xing and Lin, Kun-Yu and Wei, Xihan and Hu, Jian-Fang and Xie, Xiaohua and Zheng, Wei-Shi},
  booktitle={Proc. ICCV},
  pages={21778--21788},
  year={2025}
}

@article{hurst2024gpt,
  title={Gpt-4o system card},
  author={Hurst, Aaron and Lerer, Adam and Goucher, Adam P and Perelman, Adam and Ramesh, Aditya and Clark, Aidan and Ostrow, AJ and Welihinda, Akila and Hayes, Alan and Radford, Alec and others},
  journal={arXiv preprint arXiv:2410.21276},
  year={2024}
}

@article{team2024gemini,
  title={Gemini 1.5: Unlocking multimodal understanding across millions of tokens of context},
  author={Team, Gemini and Georgiev, Petko and Lei, Ving Ian and Burnell, Ryan and Bai, Libin and Gulati, Anmol and Tanzer, Garrett and Vincent, Damien and Pan, Zhufeng and Wang, Shibo and others},
  journal={arXiv preprint arXiv:2403.05530},
  year={2024}
}

@article{wang2026mossvl,
  title={{MOSS-VL} Technical Report},
  author={Wang, Pengyu and Tan, Chenkun and Zhou, Shaojun and Zhou, Qirui and Chen, Yanxin and He, Xingyang and Zeng, Huazheng and Cheng, Jijun and Wang, Chenghao and Qian, Xiaomeng and others},
  journal={arXiv preprint arXiv:2608.15045},
  year={2026}
}

@article{cui2026minicpmo45,
  title={{MiniCPM-o 4.5}: Towards Real-Time Full-Duplex Omni-Modal Interaction},
  author={Cui, Junbo and Xu, Bokai and Wang, Chongyi and Yu, Tianyu and Sun, Weiyue and Xu, Yingjing and Wang, Tianran and He, Zhihui and Ma, Wenshuo and Cai, Tianchi and others},
  journal={arXiv preprint arXiv:2604.27393},
  year={2026}
}

@article{qwen2025qwen25omni,
  title={{Qwen2.5-Omni} Technical Report},
  author={{Qwen Team}},
  journal={arXiv preprint arXiv:2503.20215},
  year={2025}
}

@article{qwen2025qwen3omni,
  title={{Qwen3-Omni} Technical Report},
  author={{Qwen Team}},
  journal={arXiv preprint arXiv:2509.17765},
  year={2025}
}

@misc{google2026geminilive,
  author={{Google}},
  title={{Gemini Live API} Overview and Capabilities},
  year={2026},
  howpublished={\url{https://ai.google.dev/gemini-api/docs/live-api}},
  note={Accessed 2026-10-21}
}

@article{bai2025qwen25vl,
  title={{Qwen2.5-VL} Technical Report},
  author={Bai, Shuai and Chen, Keqin and Liu, Xuejing and Wang, Jialin and Ge, Wenbin and Song, Sibo and Dang, Kai and Wang, Peng and Wang, Shijie and Tang, Jun and others},
  journal={arXiv preprint arXiv:2502.13923},
  year={2025}
}

@article{yan2026proactvl,
  title={{Proact-VL}: A Proactive {VideoLLM} for Real-Time {AI} Companions},
  author={Yan, Weicai and Dai, Yuhong and Ran, Qi and Li, Haodong and Lin, Wang and Jin, Tao and Xie, Xing and Liao, Hao and Lian, Jianxun},
  journal={arXiv preprint arXiv:2603.03447},
  year={2026}
}

@article{zhang2026streamarena,
  title={{StreamArena}: Toward Continuous, Interactive, and Long-Horizon Agentic Streaming Video Understanding},
  author={Zhang, Xichen and Li, Guankai and Zhu, Yinghao and Wang, Shijian and Wu, Sitong and Yu, Shaozuo and Chu, Meng and Lu, Yuan and Jia, Jiaya},
  journal={arXiv preprint arXiv:2608.05703},
  year={2026}
}

@misc{qwen2026qwen35,
  author={{Qwen Team}},
  title={{Qwen3.5-9B} Model Card},
  year={2026},
  howpublished={\url{https://huggingface.co/Qwen/Qwen3.5-9B}},
  note={Accessed 2026-10-23}
}

@misc{qwen2026qwen38,
  author={{Qwen Team}},
  title={{Qwen3.8-Max}: A New Bar for Coding and Cowork},
  year={2026},
  month=aug,
  howpublished={\url{https://qwen.ai/blog?id=qwen3.8}},
  note={Accessed 2026-10-23}
}

@article{qwen2026qwen38omni,
  author={{Qwen Team}},
  title={{Qwen3.8-Omni}: Towards Native Omni-Modal Agents},
  journal={arXiv preprint arXiv:2609.25611},
  year={2026}
}

@misc{google2026gemini38flash,
  author={{Google}},
  title={{Gemini 3.8 Flash}},
  year={2026},
  howpublished={\url{https://ai.google.dev/gemini-api/docs/models/gemini-3.8-flash}},
  note={Accessed 2026-10-23}
}

@misc{google2026gemini38live,
  author={{Google}},
  title={{Gemini 3.8 Live}},
  year={2026},
  howpublished={\url{https://ai.google.dev/gemini-api/docs/models/gemini-3.8-live}},
  note={Accessed 2026-10-23}
}

@misc{google2026geminiproactivity,
  author={{Google}},
  title={{Gemini Live API} Capabilities Guide},
  year={2026},
  howpublished={\url{https://ai.google.dev/gemini-api/docs/live-api/capabilities}},
  note={Accessed 2026-10-23}
}

@misc{bytedance2026seed21,
  author={{ByteDance Seed}},
  title={{Seed2.1}},
  year={2026},
  howpublished={\url{https://seed.bytedance.com/en/seed2_1}},
  note={Accessed 2026-10-23}
}

@misc{bytedance2026seedrealtime,
  author={{ByteDance Seed}},
  title={{SeedRealtime} Audio-Visual Full-Duplex {LLM} Released: Toward Omni-Modal Natural Interaction},
  year={2026},
  month=aug,
  howpublished={\url{https://seed.bytedance.com/en/blog/seedrealtime-audio-visual-full-duplex-llm-released-toward-omni-modal-natural-interaction}},
  note={Accessed 2026-10-23}
}

@inproceedings{lu-etal-2026-omniinteract,
  title     = {{OmniInteract}: Benchmarking Real-World Streaming Interaction for Real-Time Omnimodal Assistants},
  author    = {Lu, Xudong and Li, Xueying and Wang, Annan and Bo, Yang and Chen, Jinpeng and Li, Zengliang and Yang, Nianzu and Liu, Rui and Yang, Xue and Hou, Jingwen and Li, Hongsheng},
  booktitle = {Proceedings of the 2026 Conference on Empirical Methods in Natural Language Processing},
  pages     = {11272--11293},
  year      = {2026},
  address   = {Budapest, Hungary},
  publisher = {Association for Computational Linguistics},
  doi       = {10.18653/2026.emnlp-main.490},
  url       = {https://aclanthology.org/2026.emnlp-main.490/}
}

@article{vllmomni2026technicalreport,
  title={{vLLM-Omni} Technical Report: A Unified Serving Runtime for Omni-Modality Generation},
  author={{vLLM-Omni Team}},
  journal={arXiv preprint arXiv:2610.09307},
  year={2026}
}

@inproceedings{ding2023mevis,
  author    = {Ding, Henghui and Liu, Chang and He, Shuting and
               Jiang, Xudong and Loy, Chen Change},
  title     = {{MeViS}: A Large-scale Benchmark for Video Segmentation
               with Motion Expressions},
  booktitle = {Proceedings of the IEEE/CVF International Conference
               on Computer Vision (ICCV)},
  pages     = {2694--2703},
  year      = {2023}
}

@inproceedings{seo2020urvos,
  author    = {Seo, Seonguk and Lee, Joon-Young and Han, Bohyung},
  title     = {{URVOS}: Unified Referring Video Object Segmentation
               Network with a Large-Scale Benchmark},
  booktitle = {Computer Vision -- ECCV 2020},
  pages     = {208--223},
  publisher = {Springer},
  year      = {2020}
}

@inproceedings{sun2022dancetrack,
  author    = {Sun, Peize and Cao, Jinkun and Jiang, Yi and
               Yuan, Zehuan and Bai, Song and Kitani, Kris and Luo, Ping},
  title     = {{DanceTrack}: Multi-Object Tracking in Uniform
               Appearance and Diverse Motion},
  booktitle = {Proceedings of the IEEE/CVF Conference on Computer
               Vision and Pattern Recognition (CVPR)},
  pages     = {20993--21002},
  year      = {2022}
}

@inproceedings{cui2023sportsmot,
  author    = {Cui, Yutao and Zeng, Chenkai and Zhao, Xiaoyu and
               Yang, Yichun and Wu, Gangshan and Wang, Limin},
  title     = {{SportsMOT}: A Large Multi-Object Tracking Dataset
               in Multiple Sports Scenes},
  booktitle = {Proceedings of the IEEE/CVF International Conference
               on Computer Vision (ICCV)},
  pages     = {9921--9931},
  year      = {2023}
}

@inproceedings{lin2014coco,
  author    = {Lin, Tsung-Yi and Maire, Michael and Belongie, Serge and
               Hays, James and Perona, Pietro and Ramanan, Deva and
               Doll{\'a}r, Piotr and Zitnick, C. Lawrence},
  title     = {{Microsoft COCO}: Common Objects in Context},
  booktitle = {Computer Vision -- ECCV 2014},
  pages     = {740--755},
  publisher = {Springer},
  year      = {2014}
}

\end{document}